\documentclass[9pt,twocolumn,twoside]{pnas-new}

\templatetype{pnasresearcharticle} 

\setboolean{displaywatermark}{false}

\usepackage{amsmath,amsfonts,bm}
\usepackage{xcolor}
\usepackage{fontawesome}

\def\eqref#1{(\ref{#1})}

\def\1{\bm{1}}

\DeclareMathAlphabet{\mathsfit}{\encodingdefault}{\sfdefault}{m}{sl}
\SetMathAlphabet{\mathsfit}{bold}{\encodingdefault}{\sfdefault}{bx}{n}

\usepackage{algorithm}
\usepackage{multirow}  
\usepackage{algpseudocode}
\usepackage{listings}
\usepackage{xcolor}
\usepackage{soul}
\usepackage{float}
\usepackage{graphicx}
\usepackage{amsmath}  
\usepackage{microtype}
\usepackage{lastpage}
\usepackage{hyperref}
\usepackage{rotating}
\usepackage{pifont}
\usepackage{bbding}
\PassOptionsToPackage{verbose=silent}{microtype}

\fancypagestyle{arxivstyle}{%
  \fancyhf{}%
  \fancyfoot[C]{\footerfont\thepage}%
}
\begin{document}

\title{Scaling Makes It Possible: How Large Models Master Impossible Languages}

\author[1]{Yaning Jia}
\author[1]{Chunhui Zhang}
\author[1]{Chiyu Ma}
\author[1]{Zhongyu Ouyang}
\author[1]{Ivory Yang}
\author[1]{Soroush Vosoughi}

\affil[1]{Dartmouth College}


\significancestatement{We investigate how language models (LMs) learn impossible languages that defy human linguistic constraints. Our findings reveal that as the model scale increases, LMs shift from syntactic reliance to semantic abstraction, enabling them to master impossible languages. Billion-scale models exhibit reduced sensitivity to syntactic perturbations and enhanced semantic understanding, challenging traditional linguistic theories prioritizing syntax. This highlights a fundamental difference between human and LM language learning mechanisms. By bridging linguistic theory and computational modeling, our study provides new insights into how LMs generalize beyond conventional language structures, reshaping our understanding of their cognitive and computational capabilities.}

\authorcontributions{
YJ and CZ designed and conducted the experiments, analyzed data, and wrote the manuscript. CM contributed to experimental design, interpreted results, and provided theoretical support. SV supervised the entire research, conceptualized the project, guided analysis and interpretation, and revised the manuscript critically for important intellectual content. All authors read, provided critical feedback, and approved the manuscript.
}
\correspondingauthor{\textsuperscript{2}To whom correspondence should be addressed. E-mail: soroush.vosoughi@dartmouth.edu}


\begin{abstract}
While language models (LMs) excel in natural language tasks, their ability to learn impossible languages—those that violate the structural, cognitive, or logical constraints that make human languages learnable and functional—remains debatable. This controversy is further fueled by a divide between linguistic theorists and empirical researchers working with small-scale LMs. To address this controversy, we introduce \textit{semantic coherence}. This concept measures how effectively semantic information is expressed in language structures and examines the learning capabilities of LMs across multiple scales. By fine-tuning billion-parameter LMs on impossible language samples, we demonstrate that larger LMs achieve similar performance on impossible and possible languages, whereas million-parameter LMs exhibit significant degradation when trained on impossible languages. This performance gap widens as semantic complexity increases. We further show that scaling up the LM size fundamentally shifts the learning focus from grammatical rules to semantic understanding, with billion-scale LMs exhibiting a threefold reduction in sensitivity to low-level linguistic perturbations while maintaining stable performance under high-level semantic perturbations. 
\end{abstract}

\dates{}
\doi{}

\maketitle
\thispagestyle{arxivstyle}
\ifthenelse{\boolean{shortarticle}}{\ifthenelse{\boolean{singlecolumn}}{\abscontentformatted}{\abscontent}}{}

\firstpage[5]{5}

\dropcap{L}anguage models (LMs) excel at various large-scale natural language tasks, such as text generation \cite{radford2019language}, machine translation \cite{vaswani2017attention}, and question answering \cite{yang2019xlnet}, demonstrating a clear proficiency in human-learnable, rule-based structures known as \textit{possible languages}. However, it remains debated whether LMs can also learn \textit{impossible languages}, which violate structural or cognitive constraints that define human language. Linguistic theorists like Chomsky argue that LMs lack mechanisms to distinguish possible from impossible languages, suggesting that both should be equally learnable \cite{chomsky2023noam, moro2023large, bolhuis2024three}. Conversely, recent empirical studies have challenged this view by systematically constructing impossible languages and showing that small-scale LMs struggle to learn them as effectively as possible languages \cite{kallini2024mission}.

These empirical studies primarily utilize small-scale models such as GPT-2 (with millions of parameters) trained from scratch to assess their learning boundaries \cite{kallini2024mission, radford2019language}. They found that GPT-2 learns possible languages more efficiently and is notably sensitive to grammatical rules, indicating a strong syntactic bias. However, we argue that such findings might not generalize to larger-scale models, which are known to better capture semantic information.

Traditional linguistic theory emphasizes grammar as essential for systematically encoding and communicating semantic content through transformational rules~\cite{chomsky2002syntactic, chomsky2014aspects}. Possible languages adhere to cognitive and grammatical constraints, enabling structured semantic encoding and efficient learning. Impossible languages violate these constraints, resulting in structures that are unnatural or challenging for human cognition.

Nevertheless, human languages with limited grammatical structure, such as \textit{Pirahã} spoken by indigenous Amazonian communities \cite{nevins2009piraha}, or symbolic systems such as Nigeria’s Nsibidi \cite{kotei1972west}, Ghana’s Adinkra symbols \cite{willis1998adinkra}, and Angola’s Lusona/Sona geometric notation \cite{wang2024comparative}, demonstrate effective semantic communication without traditional grammar. This suggests that semantic information alone, irrespective of syntactic complexity, can facilitate language acquisition. To quantify this idea, we introduce \textbf{semantic coherence}, defined as the ease with which semantic information is conveyed within a linguistic structure. We hypothesize that larger-scale LMs are better equipped to comprehend semantic coherence, thus enabling more efficient learning of impossible languages.

Interestingly, our findings at larger scales diverge from recent empirical results with smaller models \cite{kallini2024mission} and align more closely with theoretical claims by linguistic theorists \cite{chomsky2023noam}. Specifically, we show: \textbf{i)} Billion-scale LMs such as Llama3.2 and Qwen2.5 significantly outperform smaller GPT-2 models in learning impossible languages, especially those with low-level perturbations. \textbf{ii)} Smaller models (e.g., GPT-2) emphasize grammatical structures, limiting their semantic understanding under syntactic perturbations. Conversely, larger models prioritize high-level semantic comprehension and thus exhibit reduced sensitivity to low-level syntactic disruptions. \textbf{iii)} Our proposed Lipschitz-Optimizer further improves LMs’ performance and convergence on impossible languages, without necessitating architectural modifications.

\begin{figure*}[t]
\begin{center}
\includegraphics[width=0.9\linewidth]{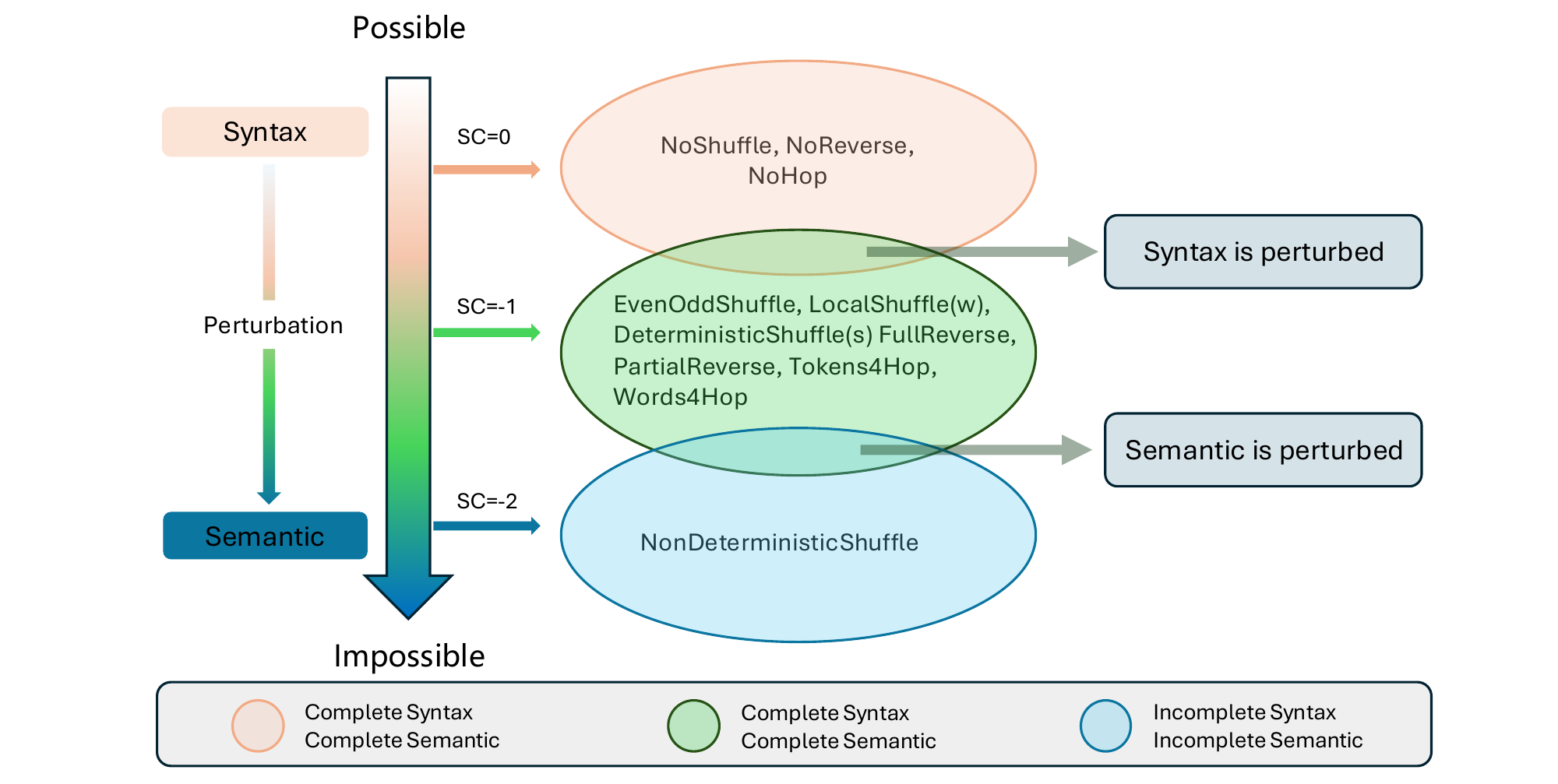}
\end{center}
\caption{Effects of language perturbations on syntactic and semantic information preservation. \textbf{SC (semantic coherence)} is categorized into three levels: SC = 0 (languages with complete syntax and semantics), SC = -1 (languages with incomplete syntax but complete semantics), and SC = -2 (languages with both incomplete syntax and semantics). Based on their perturbation functions, impossible languages are classified into three types: \textbf{Shuffle Languages} (\texttt{EvenOddShuffle}, \texttt{LocalShuffle(w)}, \texttt{DeterministicShuffle(s)}, \texttt{FullReverse}, \texttt{PartialReverse}, \texttt{Tokens4Hop}, \texttt{Words4Hop}), \textbf{Reverse Languages} (\texttt{FullReverse} and \texttt{PartialReverse}), and \textbf{Hop Languages} (\texttt{Tokens4Hop} and \texttt{Words4Hop}). For control groups, \texttt{NoShuffle}, \texttt{NoReverse}, and \texttt{NoHop} represent natural languages corresponding to each class of impossible languages.}
\vspace{0.2cm}
\label{fig:languages-classification}
\end{figure*} 

Overall, we argue that large LMs' capacity to learn impossible languages primarily stems from their enhanced ability to comprehend and utilize semantic information. \textbf{When semantic content is sufficiently coherent, large-scale LMs can effectively master languages irrespective of their syntactic complexity (or even impossibility).}

\begin{table}[h]
\caption{Impact of language perturbations on linguistic information preservation. The table categorizes perturbations by Perturbation Level (PL: High or Low) and Semantic Coherence (SC: 0, -1, -2). A checkmark ($\checkmark$) indicates complete retention of semantic or syntactic information, while a cross (\ding{55}) indicates incomplete retention. See Appendix A for more cases about different perturbations.}
\centering
\footnotesize
\setlength{\tabcolsep}{6pt}
\renewcommand{\arraystretch}{1.1}
\begin{tabular}{llccr}
\toprule
\textbf{PL} & 
\textbf{Perturbation} & \textbf{Semantic} & \textbf{Syntactic} & \textbf{SC} \\
\midrule
\multirow{5}{*}{{High}} 
& \texttt{NoShuffle}             & $\checkmark$  & $\checkmark$  & $0$ \\
& \texttt{EvenOddShuffle}       & $\checkmark$  & \ding{55}      & $-1$ \\
& \texttt{LocalShuffle(w)}      & $\checkmark$  & \ding{55}      & $-1$ \\
& \texttt{DeterministicShuffle(s)}  & $\checkmark$  & \ding{55}  & $-1$ \\
& \texttt{NonDeterministicShuffle}      & \ding{55}      & \ding{55}     & $-2$ \\
\midrule
\multirow{6}{*}{{Low}}
& \texttt{NoReverse}            & $\checkmark$  & $\checkmark$  & $0$ \\
& \texttt{FullReverse}          & $\checkmark$  & \ding{55}      & $-1$ \\
& \texttt{PartialReverse}       & $\checkmark$  & \ding{55}      & $-1$ \\
& \texttt{NoHop}                & $\checkmark$  & $\checkmark$  & $0$ \\
& \texttt{Tokens4Hop}          & $\checkmark$  & \ding{55}      & $-1$ \\
& \texttt{Words4Hop}           & $\checkmark$  & \ding{55}      & $-1$ \\
\bottomrule
\end{tabular}
\label{tab:perturbation-class}
\end{table}

\begin{figure*}[t]
\begin{center}
\includegraphics[width=0.9\linewidth]{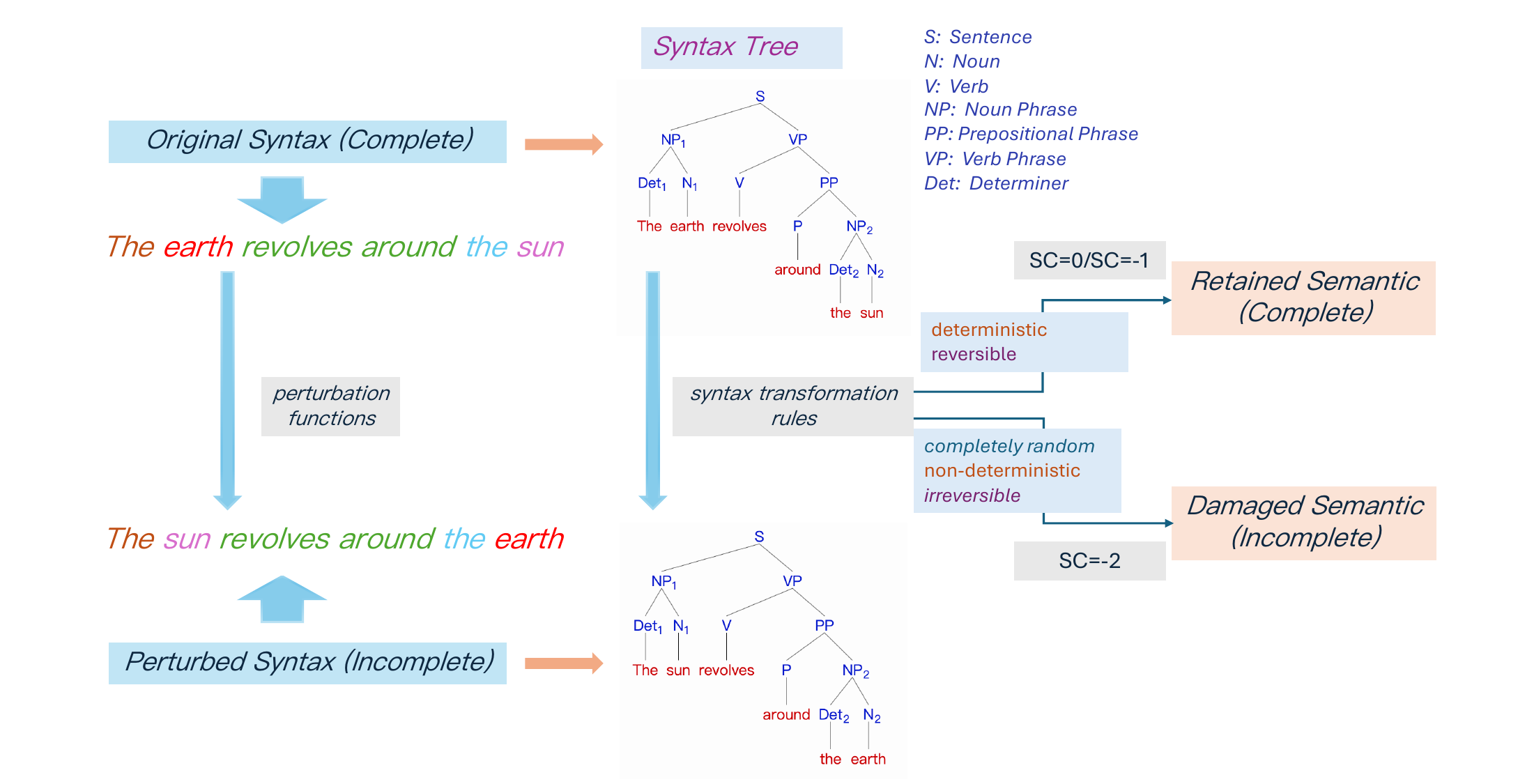}
\end{center}
\caption{Illustration of complete and incomplete syntax and semantic information retention. This figure presents a hierarchical flow demonstrating how perturbation functions impact language structures and influence their learnability. On the left, an example sentence with complete syntax retention, \textit{The earth revolves around the sun}, is shown. When perturbation functions are applied, the sentence undergoes syntactic transformations, resulting in an altered structure, \textit{The sun revolves around the earth}, which leads to incomplete syntax retention. The middle section illustrates changes in syntax tree structures under perturbations, highlighting how these transformations contribute to syntactic incompleteness. The right side categorizes semantic retention into two possible outcomes: 
\textbf{complete semantic retention} occurs when syntax transformation rules are completely random, non-deterministic, and irreversible, preventing LMs from learning these rules. In contrast, \textbf{incomplete semantic retention} arises when these rules are deterministic and reversible, allowing LMs to learn them. Additionally, the figure provides key linguistic components, including \textbf{S} (Sentence), \textbf{N} (Noun), \textbf{V} (Verb), \textbf{NP} (Noun Phrase), \textbf{PP} (Prepositional Phrase), \textbf{VP} (Verb Phrase), and \textbf{Det} (Determiner), as references for interpreting the syntax tree structures.}

\vspace{0.2cm}
\label{fig:syntax-semantic}
\end{figure*}

\section*{Results}
To explore LMs' capability boundary of learning possible and impossible languages, we examine LMs across different scales and languages with different possibilities. We construct a series of impossible languages based on the impossibility continuum proposed by Mission~\cite{kallini2024mission} and classify them into different categories based on the preservation of linguistic information, as shown in Table~\ref{tab:perturbation-class} and Figure~\ref{fig:languages-classification}. The table evaluates each type of impossible language in terms of semantic retention, syntactic structure, and overall coherence. To illustrate this classification, we provide a detailed explanation in Figure~\ref{fig:syntax-semantic}.

For semantic retention, applying perturbation functions to natural languages alters their syntactic structure, meaning the perturbed languages exhibit incomplete syntactic retention. However, suppose LMs can learn the syntactic transformation rules. In that case, the perturbed language effectively becomes a structured transformation of the original, allowing it to retain complete semantic information and remain semantically intact. Among the various perturbations, only \texttt{NonDeterministicShuffle} is entirely random, making it impossible for LMs to learn any consistent transformation rule. As a result, the corresponding perturbed language fails to retain complete semantic meaning. For instance, given the perturbed sentence \textit{The sun revolves around the earth}, LMs cannot reliably recover its intended meaning, \textit{The earth revolves around the sun}, rendering this type of language semantically incomplete.

In our experiments, we choose GPT-2~\cite{radford2019language}, Llama3.2~\cite{grattafiori2024llama, liu2024spinquant}, and Qwen2.5~\cite{yang2024qwen2} language models and train them on the original BabyLM dataset~\cite{warstadt2023call}. Through these experiments, we validate LLMs' learnability from four different views: \textbf{(1)} Different impossible languages are learnable if they retain semantics; \textbf{(2)} Scaling LM makes learning impossible languages possible; \textbf{(3)} Scaling reduce LMs' sensitivity to structural linguistic Features;  \textbf{(4)}~Lipschitz-Optimizer enhances LMs' learnability of impossible Languages.

\subsection{Different Impossible Languages are All Learnable}

\begin{figure*}[h]
\begin{center}
\includegraphics[width=1\linewidth]{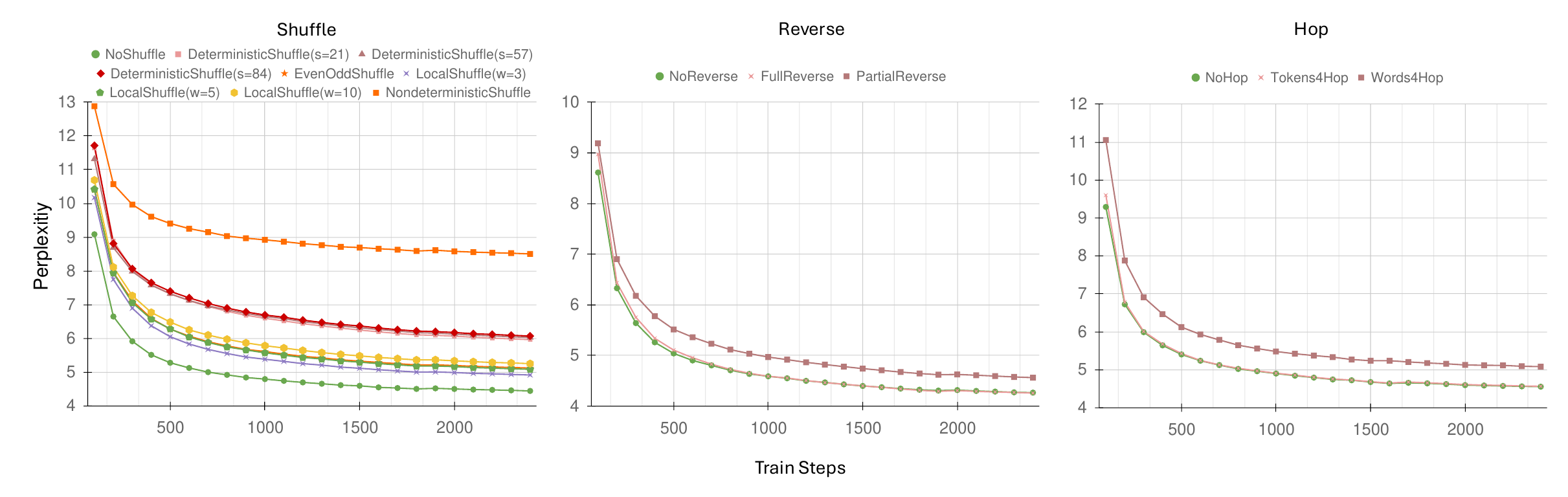}
\end{center}
\caption{Perplexities vary with training steps across different impossible languages on the test data. Lower perplexities indicate a better understanding of these languages by the language models. From left to right, perturbation functions are \textbf{Shuffle}, \textbf{Reverse}, and \textbf{Hop}, respectively. \texttt{NoShuffle}, \texttt{NoRverse}, and \texttt{NoHop} are controls, denoting unperturbed languages. All results are based on \textbf{Llama3.2-3B}.}
\vspace{0.2cm}
\label{fig:exp1-perplexity}
\end{figure*}
Our study demonstrates that LLMs can learn impossible languages altered by structural changes, artificial variants of natural language perturbed by systematic transformations. By fine-tuning Llama3.2-3B on these variants, we show that semantic coherence governs LLM's ability to master impossible languages. Our experiments reveal that despite significant syntactic alterations, LLM successfully learns these impossible languages when semantic information remains complete, indicating that semantic coherence outweighs syntactic fidelity in language learning processes.

Impossible languages are categorized into \textbf{shuffle}, \textbf{reverse}, and \textbf{hop} types based on their perturbation functions, as detailed in \hyperref[material-and-method]{Materials and Methods}. These different impossible languages present varying degrees of semantic coherence, which correlates with their difficulty for language models. We measure the learnability of LLMs for these impossible languages by their perplexity on corresponding test datasets (Figure~\ref{fig:exp1-perplexity}). Our analysis reveals two key phenomena: \textbf{(a)} The learnability varies significantly across impossible language types, with shuffle languages generally exhibiting the highest perplexities (most of them are around 6), followed by hop languages (perplexities approach 5 or lower), and then reverse languages (perplexities are all below~5);
\textbf{(b)}~Remarkably, LLMs learn most of these languages effectively (perplexities are all below~5), except for \texttt{NondeterministicShuffle} languages (perplexity approaches~9). Even more surprisingly, for \texttt{FullReverse} and \texttt{Tokens4Hop} variants, LLMs demonstrate better learning performance (lower perplexity) than unperturbed languages.

From these phenomena, we conclude that semantic coherence determines the difficulty of learning languages, while semantic integrity governs whether languages can be mastered as efficiently as natural languages. Regarding phenomenon \textbf{(a)}, the varying degrees of semantic coherence across language types directly correlate with their learnability: shuffle languages, having the lowest semantic coherence, are the most challenging to learn, while reverse and hop languages with higher semantic coherence are comparatively easier to master. For phenomenon \textbf{(b)}, \texttt{NondeterministicShuffle} languages lack complete semantic information, making them the most difficult to learn. Conversely, \texttt{FullReverse} and \texttt{Tokens4Hop} languages maintain high semantic coherence and complete semantic information, enabling LLMs to learn these languages efficiently—even more efficiently than natural languages in some cases.

\subsection{Scaling LM Makes Learning Possible}

Previous research has established that LM performance scales predictably with parameter count \cite{kaplan2020scaling}. Building on this foundation, our findings demonstrate that model scale is also a critical determinant of an LM's capacity to learn impossible languages, extending the scaling law paradigm to these structurally perturbed language variants. As LM scale increases, we observe two key phenomena: first, the capability to learn impossible languages improves dramatically; second, the performance gap between impossible and natural languages progressively narrows. This scaling effect reveals that larger LMs can effectively master various impossible languages, transforming what was previously considered unlearnable into achievable learning tasks. 

The differences in LM scales—spanning smaller architectures to larger ones—are quantified across three dimensions: parameter counts, training token volumes, and computational costs (measured in ZettaFLOPs), as summarized in Table~\ref{tab:model-scale}. This systematic comparison demonstrates how impossible language learning capabilities scale predictably with model size, revealing a fundamental relationship between scale and linguistic adaptability. To investigate the relationship between model scale and impossible language learning, we employed a diverse range of LMs, including GPT-2 (117M), Llama3.2 (1B-3B), and Qwen2.5 (0.5B-14B). Table~\ref{tab:perturbation_perplexity_gpt_llama} and Table~\ref{tab:perturbation_perplexity_qwen} reveal two key phenomena: \textbf{(a)} shuffle languages consistently exhibit higher perplexities (10\%--60\% higher) and substantially larger perplexity gaps (5-15 $time$ greater) compared to reverse and hop languages, confirming their greater learning difficulty; \textbf{(b)} as model scale increases, perplexities decrease substantially, with particularly dramatic improvements observed when scaling from GPT-2 (117M) to Llama3.2 (1B), where perplexities on Llama3.2 (1B) are approximately 1/5 of those on GPT-2 (117M). Additionally, perplexity gaps approach zero, especially for reverse languages and hop languages, with their perplexity gaps on Qwen2.5-14B falling below 3\%; however, for all LMs, perplexity gaps of \texttt{NondeterministicShuffle} languages increase gradually as scale increases. Generally, these findings suggest that larger LMs not only learn impossible languages more effectively but also become increasingly insensitive to linguistic perturbations.
 
Phenomenon \textbf{(a)} can be attributed to shuffling languages having reduced semantic coherence and more severely disrupted syntactic structures, further validating our hypothesis that semantic coherence determines language learning difficulty. Regarding phenomenon \textbf{(b)}, scaling substantially enhances LMs' ability to learn impossible languages by enabling larger models to transcend syntactic perturbations and access the underlying semantic information preserved in these languages. The abnormal trend for \texttt{NondeterministicShuffle} languages occurs because these languages are semantically incomplete, making it increasingly difficult for LMs to capture sufficient semantic information compared to natural language as the model scale increases. Consequently, as LM scale increases, the learnability gap between impossible and natural languages diminishes, particularly for impossible languages that maintain complete semantic information despite their syntactic alterations.

\begin{table*}[htbp]
\caption{Perplexity varies under input perturbations across GPT-2 and Llama3.2 model scales. Values show perplexities on each perturbation; the percentage in brackets indicates the \textit{perplexity gap}, which is the percentage increase in the model's perplexity on the impossible language and the corresponding natural language. \texttt{NoShuffle}, \texttt{NoRverse}, and \texttt{NoHop} are controls, as baseline. A smaller gap indicates a better understanding of impossible languages.}
\centering
\small
\begin{tabular}{@{}p{5.5cm}p{2.5cm}p{2.5cm}p{2.5cm}@{}}
\toprule
\textbf{Perturbation} & \textbf{GPT-2 (114M)} & \textbf{Llama3.2 (1B)} & \textbf{Llama3.2 (3B)} \\ 
\midrule
\texttt{NoShuffle(Baseline)} & 23.24 & 4.58 & 4.40 \\
\texttt{EvenOddShuffle} & 30.28\,(+30.3\%) & 5.36\,(+17.0\%) & 5.06\,(+15.0\%) \\
\texttt{DeterministicShuffle} & 36.79\,(+58.3\%) & 6.38\,(+39.3\%) & 6.01\,(+36.6\%) \\
\texttt{NondeterministicShuffle} & 42.73\,(+83.9\%) & 8.62\,(+88.2\%) & 8.44\,(+91.8\%) \\
\addlinespace[0.3em]
\midrule
\texttt{NoReverse(Baseline)} & 21.00 & 4.36 & 4.20 \\
\texttt{FullReverse} & 21.24\,(+1.1\%) & 4.34\,(-0.5\%) & 4.18\,(-0.5\%) \\
\texttt{PartialReverse} & 23.08\,(+9.9\%) & 4.69\,(+7.6\%) & 4.49\,(+6.9\%) \\
\addlinespace[0.3em]
\midrule
\texttt{NoHop(Baseline)} & 24.26 & 4.69 & 4.48 \\
\texttt{Tokens4Hop} & 26.22\,(+8.1\%) & 4.71\,(+0.4\%) & 4.49\,(+0.2\%) \\
\texttt{Words4Hop} & 32.14\,(+32.9\%) & 5.37\,(+14.5\%) & 5.07\,(+13.2\%) \\
\bottomrule
\end{tabular}

\label{tab:perturbation_perplexity_gpt_llama}
\end{table*}

\begin{table*}[h]
\caption{Perplexity varies under input perturbations across Qwen2.5 model scales. Values show perplexities on each perturbation; the percentage in brackets indicates the \textit{perplexity gap}, which is the percentage increase in the model's perplexity on the impossible language and the corresponding natural language. \texttt{NoShuffle}, \texttt{NoRverse}, and \texttt{NoHop} are controls, as baseline. A smaller gap indicates a better understanding of impossible languages.}
\centering
\small
\begin{tabular}{@{}llllll@{}}
\toprule
\textbf{Perturbation} & \textbf{Qwen2.5 (0.5B)} & \textbf{Qwen2.5 (1.5B)} & \textbf{Qwen2.5 (3B)} & \textbf{Qwen2.5 (7B)} & \textbf{Qwen2.5 (14B)} \\ 
\midrule
\texttt{NoShuffle(Baseline)} & 2.71 & 2.64 & 2.59 & 2.49 & 2.47 \\
\texttt{EvenOddShuffle} & 3.07\,(+13.3\%) & 2.98\,(+12.9\%) & 2.92\,(+12.7\%) & 2.79\,(+12.1\%) & 2.72\,(+10.1\%) \\
\texttt{DeterministicShuffle} & 3.30\,(+21.8\%) & 3.23\,(+22.3\%) & 3.17\,(+22.4\%) & 3.05\,(+22.5\%) & 2.89\,(+17.0\%) \\
\texttt{NondeterministicShuffle} & 3.62\,(+33.6\%) & 3.57\,(+35.2\%) & 3.54\,(+36.7\%) & 3.47\,(+39.4\%) & 3.44\,(+39.3\%) \\
\addlinespace[0.3em]
\midrule
\texttt{NoReverse(Baseline)} & 2.55 & 2.49 & 2.44 & 2.36 & 2.33 \\
\texttt{PartialReverse} & 2.69\,(+5.5\%) & 2.62\,(+5.2\%) & 2.57\,(+5.3\%) & 2.47\,(+4.7\%) & 2.40\,(+2.9\%) \\
\addlinespace[0.3em]
\midrule
\texttt{NoHop(Baseline)} & 2.76 & 2.68 & 2.63 & 2.51 & 2.47 \\
\texttt{Words4Hop} & 2.82\,(+2.2\%) & 2.74\,(+2.2\%) & 2.68\,(+1.9\%) & 2.56\,(+2.0\%) & 2.50\,(+1.2\%) \\
\bottomrule
\end{tabular}

\label{tab:perturbation_perplexity_qwen}
\end{table*}

\subsection{Scaling Reduce Structural Linguistic Sensitivity}

\begin{figure*}[!h]
\begin{center}
\includegraphics[width=1\linewidth]{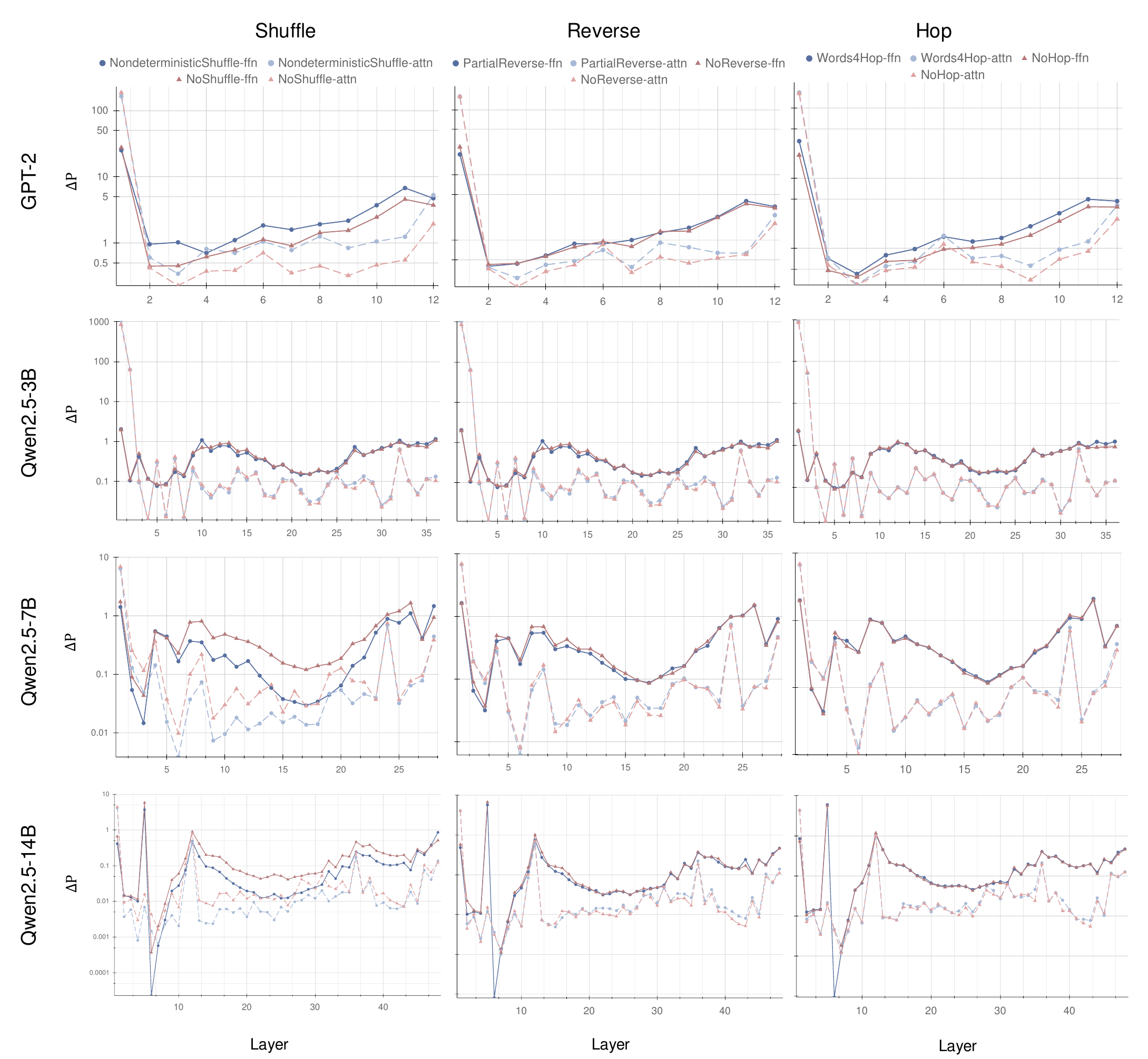}
\end{center}
\caption{Perplexity shift ($\Delta P$) measures a layer's contribution to language processing, defined as $\Delta P = P_{1} - P_{2}$, where $P_{2}$ is the perplexity before masking and $P_{1}$ is the perplexity after masking. A higher $\Delta P$ indicates greater importance. This figure shows $\Delta P$ across layers under layer-wise masking for three types of impossible languages (\texttt{NonterministicShuffle}, \texttt{PartialReverse}, and \texttt{Words4Hop}, shown in blue) and their corresponding control groups (\texttt{NoShuffle}, \texttt{NoReverse}, and \texttt{NoHop}, shown in red). Results are presented for four model scales: \textbf{GPT-2}, \textbf{Qwen2.5-3B}, \textbf{Qwen2.5-7B}, and \textbf{Qwen2.5-14B} (top to bottom).  Bold lines represent $\Delta P$ for feed-forward network (FFN) module masking, while dashed lines indicate $\Delta P$ for attention network (ATTN) module masking.}
\vspace{0.2cm}
\label{fig:mask-layer}
\end{figure*}

\begin{figure*}[!h]
\begin{center}
\includegraphics[width=1\linewidth]{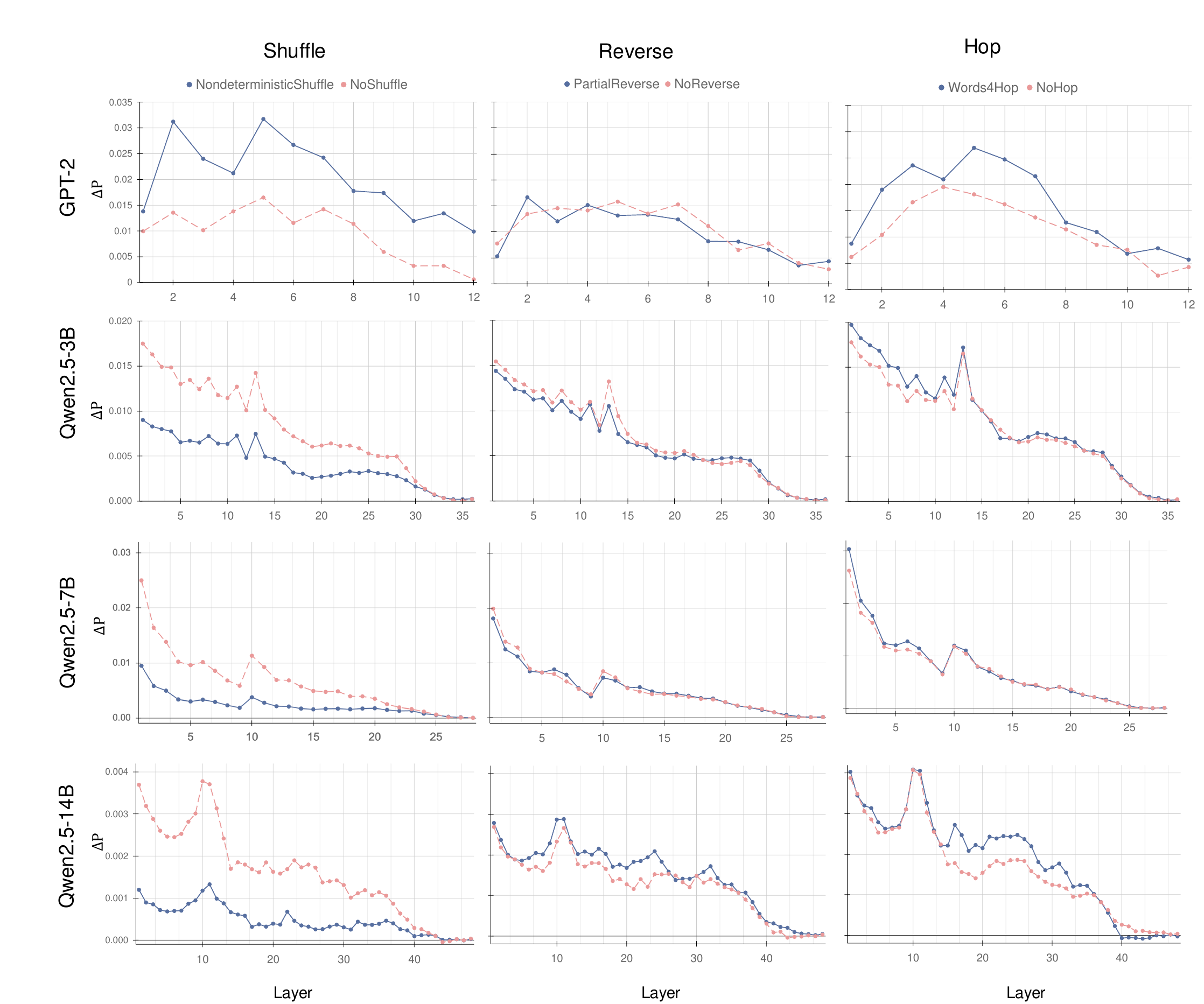}
\end{center}
\caption{Perplexity shift ($\Delta P$) measures a layer's sensitivity to noise when processing languages, 
defined as $\Delta P = P_{1} - P_{2}$, where $P_{2}$ is the model's perplexity before noise injection 
and $P_{1}$ is the perplexity afterwards. A lower $\Delta P$ indicates greater stability. The figure shows $\Delta P$ across layers under layer-wise noise injection for impossible languages (\texttt{NonterministicShuffle}, \texttt{PartialReverse}, and \texttt{Words4Hop}, shown in blue bold lines) and their corresponding control groups (\texttt{NoShuffle}, \texttt{NoReverse}, and \texttt{NoHop}, shown in light reddish brown dashed lines). Results are presented for four model scales: \textbf{GPT-2}, \textbf{Qwen2.5-3B}, \textbf{Qwen2.5-7B}, and \textbf{Qwen2.5-14B} (top to bottom).}
\vspace{0.2cm}
\label{fig:noise-layer}
\end{figure*}

Our analysis demonstrates that larger LMs exhibit superior capability in learning impossible languages. To elucidate this phenomenon, we dissect their language learning mechanisms through three key findings: \textbf{(a)} \textbf{Structural-to-semantic shift}: Scaling reduces LMs’ reliance on structural linguistic features, prioritizing semantic abstraction in impossible language processing; \textbf{(b)} \textbf{Layer-specific specialization}: Lower layers primarily capture syntactic structures, whereas higher layers emphasize semantic abstraction, aligning with findings from previous studies \cite{jawahar2019does, tenney2019you, raffel2020exploring}. Scaling amplifies higher layers’ semantic encoding capacity; \textbf{(c)} \textbf{Generalization resilience}: Enhanced semantic processing enables robust performance across both natural and impossible languages, enhancing their resilience to linguistic perturbations. 


To investigate these findings, we conduct two complementary experiments: \textbf{(1)} layer-wise masking to probe semantic representations and \textbf{(2)} layer-wise noise injection to assess syntactic and grammatical encoding.

Figure~\ref{fig:mask-layer} reveals several key phenomena: \textbf{(i)} Masking FFN layers yields significantly higher $\Delta P$ values (2-10$\times$ greater) than masking ATTN modules across perturbations. Notably, scaling reduces the $\Delta P$ gap between FFN and ATTN: on smaller LMs, the gap narrows progressively with model size, collapsing to $<0.5$ for all languages at the 14B scale. This demonstrates that FFN components dominate over attention mechanisms in impossible language acquisition, while scaling attenuates their functional divergence. \textbf{(ii)} Masking higher layers generally yields 2-3$\times$ higher $\Delta P$ values than masking lower layers (except Qwen2.5-14B showing the inverse trend), demonstrating that higher layers are more critical for LMs to master \textit{impossible languages}. \textbf{(iii)} As the model scale increases, the values of the lower layer $\Delta P$ stabilize ($\approx 1$) while the higher layer $\Delta P$ decreases sharply ($7 \to 0.5$). Currently, the $\Delta P$ gaps between the impossible and natural languages decrease - consistent with previous findings - indicating that scaling reduces the sensitivity to semantic perturbations, yet preserves the syntactic sensitivity and enhances the learnability of the LMs of impossible languages.

Figure~\ref{fig:noise-layer} demonstrates a complementary phenomenon: \textbf{(iv)}~$\Delta P$ values monotonically decrease with layer depth, with lower-layer $\Delta P$ consistently surpassing higher-layer values. This demonstrates stronger noise sensitivity in lower layers, combined with \textbf{(c)}, further validating findings \textbf{(a)} and~\textbf{(b)}. \textbf{(v)} For low-level perturbations (reverse/hop), $\Delta P$ gaps between \textit{impossible} and natural languages approach zero in Qwen models (e.g., $<0.001$ for Qwen-14B). For high-level perturbations, $\Delta P$ gaps remain pronounced but diminish with scaling across all layers (maximum $\Delta P$: 0.03→0.004 on shuffled languages), indicating that scaling reduces divergence between linguistic types and enhances LMs' overall stability for structural perturbations.

The phenomenon \textbf{(i)} can be attributed to the fundamental roles of different types of modules: FFN layers are primarily responsible for encoding syntactic and semantic properties~\cite{geva2020transformer, clark2019does, hewitt2019structural, tenney2019bert, voita2019bottom}, while attention modules primarily facilitate token selection and relationship modeling~\cite{clark2019does, NIPS2017_3f5ee243, voita2019analyzing, rogers2021primer}. This functional distinction makes FFN layers substantially more critical for learning impossible languages. Additionally, as models scale, these FFN modules develop enhanced capabilities for capturing both syntactic and semantic information, resulting in greater resilience to information absence during masking experiments. Regarding phenomenon \textbf{(ii)}, lower layers predominantly capture syntactic structure, whereas higher layers focus on semantic modeling \cite{jawahar2019does, tenney2019you, raffel2020exploring}, and semantic information proves more crucial for language acquisition. For phenomenon \textbf{(iii)}, scaling enhances LMs' ability to capture semantic information while producing relatively minor improvements in syntactic and grammatical processing. This indicates that larger LMs' improved language understanding stems primarily from enhanced semantic processing rather than syntactic comprehension. Phenomenon \textbf{(iv)} occurs because syntactic and structural features are more vulnerable to noise than semantic information; since lower layers are responsible for encoding these syntactic features, they exhibit greater sensitivity to noise perturbations. Finally, phenomenon \textbf{(v)} demonstrates that scaling enhances LMs' language understanding by improving their semantic processing capabilities while reducing reliance on syntactic information, thereby making them less sensitive to noise. Overall, as the model scale increases, LMs' semantic processing abilities improve substantially, rendering them more robust to syntactic perturbations such as noise injection.

\subsection{Lipschitz-Optimizer Enhance Learnability of Impossible Languages}
We find that by adjusting the optimizers, even without changing the model design itself, LM convergence is slightly accelerated, and stability to perturbed languages is improved, ultimately improving the learnability of LMs on impossible languages. 

\begin{figure*}[h!]
\begin{center}
\includegraphics[width=1\linewidth]{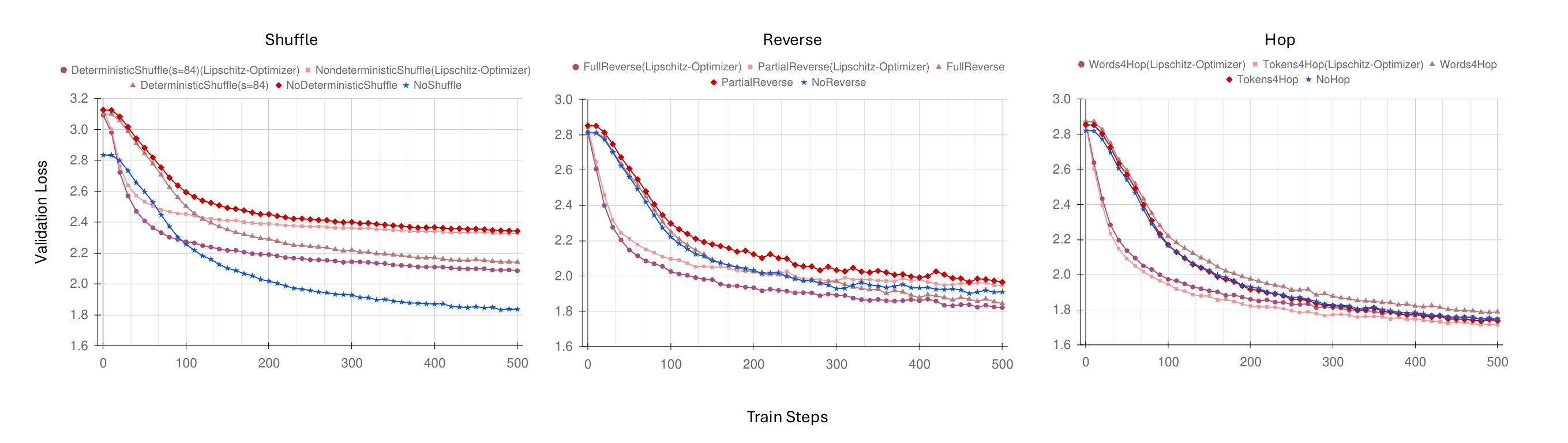}
\end{center}
\caption{Validation losses on \textbf{Llama3.2-3B} across training steps for impossible languages (\textbf{Shuffle}, \textbf{Reverse}, and \textbf{Hop}, from left to right). Results compare training with the Lipschitz Optimizer to those using the original optimizer.}
\vspace{0.2cm}
\label{fig:exp3-validloss}
\end{figure*}

The Lipschitz constant, defined as the maximum change in output space corresponding to a variation in input space for neural networks, has been widely applied across diverse domains. These applications include closed-loop system stability analysis \cite{fabiani2022neural, sun2024achieving}, stability verification \cite{huster2019limitations}, fairness guarantees~\cite{jia2024aligning}, and stability metrics for various architectures~\cite{yang2022tableformer, verine2023expressivity, gupta2022multivariate, jia2023enhancing, gama2019stability}. Additional applications extend to reachability analysis \cite{ruan2018reachability}, guaranteed generalization bounds~\cite{sokolic2017robust}, and invertible neural networks \cite{behrmann2019invertible}. By constraining the Lipschitz constant of parameters during optimization, we can enhance parameter convergence speed and improve LMs' stability when processing impossible languages. Table \ref{tab2:perplexity with Lipschitz-Optimizer} and Figure~\ref{fig:exp3-validloss} reveal two significant phenomena: \textbf{(a)} \textbf{Lipschitz-Optimizer} dramatically accelerates convergence, achieving in 200 steps what the original optimizer requires 500 steps to accomplish; \textbf{(b)} \textbf{Lipschitz-Optimizer} reduces both losses and perplexities by 3\%-5\% across all impossible language types, indicating enhanced language comprehension capabilities. Notably, for languages with lower-level perturbations (reverse and hop variants), LMs optimized with our approach outperform their performance in natural languages, demonstrating exceptional versatility and stability across diverse linguistic challenges.

The accelerated convergence observed in phenomenon \textbf{(a)} can be attributed to the Lipschitz constraint that promotes parameter smoothness, which streamlines the optimization trajectory during training. Regarding phenomenon \textbf{(b)}, \textbf{Lipschitz-Optimizer} effectively reduces LMs' sensitivity to syntactic perturbations while enhancing their ability to extract preserved semantic information from impossible languages. This dual mechanism enables LMs to develop more robust representations of linguistic content despite structural alterations, ultimately improving comprehension of impossible languages across various perturbation types. 

\begin{table*}[htbp]
\centering
\caption{Perplexities for Shuffle, Reverse, and Hop languages. \texttt{(+Lip)} denotes training process with \textbf{Lipschitz-Optimizer} on perturbed languages. \textbf{Bold} values indicate the Perplexity with \textbf{Lipschitz-Optimizer}.}
\resizebox{0.95\textwidth}{!}{
\begin{tabular}{lclclc}
\toprule
\textbf{Shuffle Setting} & \textbf{Perplexity} & \textbf{Reverse Setting} & \textbf{Perplexity} & \textbf{Hop Setting} & \textbf{Perplexity} \\
\midrule
\texttt{NoShuffle} & 5.26 & \texttt{NoReverse} & 5.00 & \texttt{NoHop} & 5.32 \\ 
\texttt{DeterministicShuffle(s=84)} & 7.37 & \texttt{PartialReverse} & 5.47 & \texttt{Tokens4Hop} & 5.35 \\ 
\texttt{NondeterministicShuffle} & 9.38 & \texttt{FullReverse} & 5.07 & \texttt{Words4Hop} & 6.11 \\ 
\texttt{DeterministicShuffle(s=84)(+Lip)} & \textbf{6.97} & \texttt{PartialReverse(+Lip)} & \textbf{5.11} & \texttt{Tokens4Hop(+Lip)} & \textbf{5.02} \\ 
\texttt{NondeterministicShuffle(+Lip)} & \textbf{9.02} & \texttt{FullReverse(+Lip)} & \textbf{4.81} & \texttt{Words4Hop(+Lip)} & \textbf{5.73} \\ 
\bottomrule
\end{tabular}%
}
\label{tab2:perplexity with Lipschitz-Optimizer}
\end{table*}

\section*{Discussion}

Our study demonstrates that LM scaling fundamentally enhances the ability to learn impossible languages through a transition from syntactic dependency to semantic abstraction. By introducing the concept of semantic coherence, which quantifies how linguistic inputs preserve meaning despite structural perturbations, we find that larger LMs more effectively extract semantic information compared to smaller models. Consequently, models such as Qwen2.5-14B achieve performance on impossible languages nearly equivalent to natural languages, provided semantic completeness remains intact. These findings, from the perspective of impossible languages, align closely with known scaling laws established on natural languages\cite{kaplan2020scaling}, showing that sufficiently large LMs develop adaptive mechanisms for semantic generalization beyond structural linguistic constraints.

Our experiments identifies three primary mechanisms facilitating this scaling effect that scaled LMs generalize effectively to impossible languages:
\textbf{(i)} \textit{structural-to-semantic shift}: Larger models increasingly depend on semantic abstraction rather than low-level syntactic features, reducing the impact of structural perturbations. Deeper layers primarily encode semantic abstractions rather than syntactic constraints, explaining why structural impossible perturbations have minimal impact on LM performance; 
\textbf{(ii)} \textit{layer-specific specialization}: FFN layers become more crucial than attention layers for semantic representation in larger LMs, gradually shifting their role from syntactic processing to semantic abstraction;
\textbf{(iii)} \textit{generalization resilience:}: larger LMs exhibit increasing resilience to linguistic perturbations, particularly in the presence of \texttt{NondeterministicShuffle} perturbations. 
While small-scale LMs are highly sensitive to syntactic disruptions, billion-scale LMs maintain stable perplexities across different impossible languages, suggesting a fundamental shift toward semantic generalization.

These findings challenge \textit{classical linguistic theories} and current perspectives on language learning. Classical linguistic theories emphasize hierarchical syntax as fundamental to human language acquisition \cite{chomsky2014aspects, berwick2016only, hauser2002faculty, pinker2003language, lidz2003infants}, categorizing languages that disrupt conventional syntactic structures as inherently unlearnable. Building upon these theoretical foundations, \textit{recent LM research} predominantly focuses on syntactic and grammatical learning mechanisms, seeking parallels between human and machine language acquisition, particularly emphasizing hierarchical syntactic structures \cite{linzen2016assessing, goldberg2019assessing, hupkes2020compositionality, warstadt2020blimp, bogdan2025emergent}. Some studies propose that explicitly pre-training LMs on syntactic and semantic constraints can improve their natural language capabilities \cite{hu2025between}. Although semantic learning has also been explored \cite{mikolov2013distributed, ilic2018deep, tenney2019bert, petroni2019language, geva2020transformer}, these investigations typically concentrate on isolated, single-model architectures, examining layers or modules independently without considering \textit{how learning strategies evolve with scaling}.

In contrast, our study uniquely emphasizes the transformative influence of scaling on language acquisition strategies. By systematically perturbing natural languages, we simultaneously evaluate syntactic and semantic learning mechanisms across a range of model sizes. Our results illustrate a fundamental shift: smaller LMs predominantly rely on syntactic rules, whereas billion-parameter LMs increasingly transcend structural constraints, prioritizing semantic abstraction. This scale-dependent semantic shift enables larger models to comprehend \textit{impossible languages}, even amid significant syntactic disruptions. Ultimately, our findings endorse the perspective that language's essential nature lies not solely in hierarchical structures, but rather in its semantic capacity to represent diverse meanings, encompassing external reality and internal cognition \cite{malik2025constructed}. These insights considerably refine our understanding of language processing in latest AI systems.

Furthermore, these results bear practical implications for the development of AI models in low-resource and synthetic language settings. If semantic information alone is sufficient for effective learning, then explicitly grammatical scaffolding may not be necessary for training robust models. This raises questions about the cruciality of formal linguistic structure in LM pretraining and whether similar shifts occur in multilingual or non-Indo-European languages. Our study elicits several open questions:
\begin{itemize}
    \item \textbf{Are LMs truly understanding the meaning of language itself?} The decreased reliance on syntactic structures suggests that models prioritize semantic coherence, but whether this corresponds to genuine understanding or statistical pattern recognition remains an open debate.
    \item \textbf{Impact on generalization to natural languages:} While our results indicate improved learnability of impossible languages, the consequences of reduced syntactic sensitivity for generalization to human-like linguistic reasoning require further investigation, especially considering that some minority languages worldwide share characteristics with impossible languages~\cite{Ghorbanpour2022impossiblelanguages, nevins2009piraha}. These languages are valuable for further systematic academic collection and research replication.
    \item \textbf{Effects on Low-Resource and Compositional Generalization:} If larger LMs rely less on structural rules, will this limit their ability to generalize in low-data regimes or for tasks requiring strong compositional reasoning?
\end{itemize}

\subsection*{Materials and Methods}
\label{material-and-method}
\paragraph{Datasets}
We conduct experiments on a series of impossible languages generated by applying various perturbation functions \cite{kallini2024mission} to the original BabyLM dataset \cite{warstadt2023call}. The BabyLM dataset is an English-language corpus containing approximately 100 million words, designed to simulate the linguistic input available to an English-speaking child. In our experiments, we utilize the text-only version of the BabyLM dataset, containing approximately 10 million words, which includes six sub-corpora: bnc\_spoken, childes, gutenberg, open\_subtitles, simple\_wiki, and switchboard.

\paragraph{Models}
\begin{table}[t] 
  \centering
  \caption{Model scaling specifications across three architectures (GPT-2, Llama-3.2, and QWen-2.5). The parameters number (unit is billion) and training tokens number (unit is billion) are referred from official reports \cite{radford2019language, yang2024qwen2, grattafiori2024llama, liu2024spinquant}, with computation measured in ZettaFLOPs. Computational costs are estimated using the heuristic FLOPs $\approx$ 6ND, where N denotes parameters and D denotes tokens~\cite{kaplan2020scaling, chowdhery2023palm}.}
  \label{tab:model-scale}
  \resizebox{1\columnwidth}{!}{%
    \begin{tabular}{lcccc}
      \toprule
      Model & Parameter (B) & Tokens (B) & ZettaFLOPs & Optimizer \\
      \midrule
      \multicolumn{5}{c}{\texttt{GPT-2}} \\
      \midrule
      117M & 0.117 & (1, 3) & 0.007 & Adam \\
      \midrule
      \multicolumn{5}{c}{\texttt{Llama-3.2}} \\
      \midrule
      1B & 1.0 & 9,000 & 54 & AdamW \\
      3B & 3.0 & 9,000 & 162 & AdamW \\
      \midrule
      \multicolumn{5}{c}{\texttt{QWen-2.5}} \\
      \midrule
      0.5B & 0.5 & 18,000 & 54 &  AdamW\\
      1.5B & 1.5 & 18,000 & 162 &  AdamW\\
      2.5B & 2.5 & 18,000 & 270 &  AdamW\\
      3.0B & 3.0 & 18,000 & 324 & AdamW \\
      7.0B & 7.0 & 18,000 & 756 & AdamW \\
      14.0B & 14.0 & 18,000 & 1,512 & AdamW \\
      \bottomrule
    \end{tabular}%
  }
\end{table}

We select GPT-2, Llama3.2, and Qwen2.5 as our experimental LMs and display the scaling trends of these LMs, as shown in Table \ref{tab:model-scale}. This comparison reveals how learning behaviors and capabilities evolve with increases in model parameters, training tokens, and computational resources. 

\paragraph{Metric}
We use perplexity as a metric for assessing language learning. Perplexity quantifies how effectively a language model predicts a set of words, being formally defined as the exponential of the entropy of the distribution. In practice, a lower perplexity score signifies a more accurate predictive model, indicating that the model is less "perplexed" by the data it encounters.

\paragraph{Impossible Languages Construction}
Impossible languages can be constructed through a series of \textit{perturbation functions} that map English input sentences to modified token sequences. By defining different perturbation functions, we create various languages spanning the impossibility continuum. These functions fall into three categories: \textbf{Shuffle}, \textbf{Reverse}, and \textbf{Hop}, each defining a distinct class of impossible languages. Shuffle languages examine how word order perturbations affect semantic and structural information preservation. Reverse languages evaluate the impact of partial or complete word order reversals on meaning capture and adaptation to non-natural linguistic structures. Hop languages analyze how long-distance dependencies and grammatical markers influence a model's ability to capture syntax and structure.

Shuffle languages, specifically, are generated by applying various reordering functions to sentence words. These are classified into five types based on the degree of structural perturbation introduced.
\begin{itemize} 
    \item \texttt{NonDeterministicShuffle}: The input sentence is tokenized and then randomly shuffled, which results in a completely unordered sequence. This process introduces full randomness, making the transformation irreversible and linguistically incoherent.
    
    \item \texttt{DeterministicShuffle(s)}: The tokenized sentence is shuffled in a deterministic manner based on the token sequence length. This transformation is reversible but removes any recognizable linguistic structure.
    
    \item \texttt{LocalShuffle(w)}: The sentence is tokenized and shuffled within fixed-size local windows of width $w$. This introduces localized perturbations in word order while maintaining some structural coherence.
    
    \item \texttt{EvenOddShuffle}: After tokenization, the sentence is reordered such that all tokens at even indices precede those at odd indices. This retains partial linear structures, but introduces long-range reordering.
    
    \item \texttt{NoShuffle}: The input sentence is tokenized, and the token sequence remains unchanged, serving as the control group.
\end{itemize}

Reverse languages are languages in which the word order within sentences is partially or completely reversed, and they are categorized into the following three types:

\begin{itemize} 
    \item \texttt{NoReverse}: The input sentence is tokenized, and a special marker \fbox{R} is inserted at a random position in the token list, serving as the control group.
    \item \texttt{PartialReverse}: Similar to \texttt{NoReverse}, but all tokens following the marker \fbox{R} are reversed. 
    \item \texttt{FullReverse}: The input sentence is tokenized, a marker \fbox{R} is inserted randomly in the token list, and the entire token sequence is reversed.
\end{itemize}

Hop languages are created based on counting rules, and these languages are classified into three types:
\begin{itemize} 
    \item \texttt{NoHop}: Add particular markers after all third-person present-tense verbs. Singular verbs are marked with \fbox{S}, and plural verbs with \fbox{P}. NOHOP retains the control group.
    \item \texttt{Tokens4Hop}: Similar to \texttt{NoHop}, but makers are placed 4 tokens after the verb.
    \item \texttt{Words4Hop}: Similar to \texttt{NoHop}, but makers are placed 4 words after the verb.
\end{itemize}

\paragraph{Lipschitz-Optimizer Implementation}
To enhance LLMs' stability when processing linguistically implausible inputs, we introduce a Lipschitz-constrained optimization framework that accelerates convergence and improves model stability. Our supplementary materials thoroughly examine LLM stability through Lipschitz continuity, establishing theoretical foundations for spectral norm regularization during training.

This framework comprises five key components: \texttt{Initial Update (init\_update)}, \texttt{Frobenius Norm Regularization (frobenius\_reg)}, \texttt{Spectral Norm Clipping (spectral\_clip)}, \texttt{Projection Radius Update (radius\_update)}, and \texttt{Weight Projection (weight\_proj)}. \texttt{init\_update} drives the model toward the loss function minimum through adaptive gradient optimization, providing reliable estimates for subsequent stability constraints. \texttt{frobenius\_reg} penalizes weight matrices' energy growth, preventing excessive parameter-space expansion, suppressing divergence in high-dimensional optimization trajectories, and enhancing generalization. \texttt{spectral\_clip} constrains the maximum gain coefficient of neural network layers by dynamically truncating singular values, ensuring controllable input-output scaling ratios and blocking gradient explosion propagation. \texttt{radius\_update} functions as a stability regulator, dynamically calibrating the feasible parameter region's radius based on real-time training feedback, balancing flexibility and stability. \texttt{weight\_proj} projects weight vectors into a controlled neighborhood anchored at initialization, preserving gradient directional information while restricting parameter drift magnitude. Detailed formula descriptions and derivations are available in the supplementary materials.

We refer to this framework as the \textbf{Lipschitz-Optimizer}, with its detailed implementation provided in Algorithm~\ref{alg:lip-opt}. It significantly enhances the stability of LLMs and accelerates their convergence speed.

\begin{minipage}{0.45\textwidth} 
    \begin{algorithm}[H]
        \caption{\footnotesize Implementation of Lipschitz-Optimizer}
        \label{alg:lip-opt}
        \begin{lstlisting}[language=python]
def lipschitz_optimizer(W_init, T, tau, mu):
    """
    Args:
        W_init: Initial weights
        T: Number of iterations
        tau: Spectral clipping threshold
        mu: Frobenius regularization factor

    Returns:
        W[T]: Optimized weights after T iterations
    """
    W = [W_init] + [None] * T
    radius = [0] * (T + 1)

    for t in range(1, T + 1):
        # Initial adaptive gradient update
        W_grad = init_update(W[t-1])

        # Frobenius norm regularization
        W_frob = frobenius_reg(W_grad, mu)

        # Spectral norm clipping
        W_clip = spectral_clip(W_frob, tau)

        # Update projection radius
        if t == 1:
            radius[t] = 1e-8
        else:
            radius[t] = radius_update(radius[t-1],
                                      W_clip, W[0])

        # Project weights towards initialization
        W_proj = weight_proj(W_clip, W[0], radius[t])

        # Final spectral clipping
        W[t] = spectral_clip(W_proj, tau)

    return W[T]
        \end{lstlisting}
    \end{algorithm}
\end{minipage}

\showacknow{} 



\bibliography{ref}

\begin{thebibliography}{10}

\bibitem{radford2019language}
A Radford, et~al., Language models are unsupervised multitask learners.
\newblock {\em\protect\JournalTitle{OpenAI blog}} (2019).

\bibitem{vaswani2017attention}
A Vaswani, Attention is all you need in {\em Advances in Neural Information
  Processing Systems}.
\newblock (2017).

\bibitem{yang2019xlnet}
Z Yang, et~al., Xlnet: Generalized autoregressive pretraining for language
  understanding in {\em Advances in Neural Information Processing Systems}.
\newblock (2019).

\bibitem{chomsky2023noam}
N Chomsky, I Roberts, J Watumull, Noam chomsky: The false promise of chatgpt.
\newblock {\em\protect\JournalTitle{The New York Times}} (2023).

\bibitem{moro2023large}
A Moro, M Greco, SF Cappa, Large languages, impossible languages and human
  brains.
\newblock {\em\protect\JournalTitle{Cortex}} (2023).

\bibitem{bolhuis2024three}
JJ Bolhuis, S Crain, S Fong, A Moro, Three reasons why ai doesn't model human
  language.
\newblock {\em\protect\JournalTitle{Nature}} (2024).

\bibitem{kallini2024mission}
J Kallini, I Papadimitriou, R Futrell, K Mahowald, C Potts, Mission: Impossible
  language models in {\em Proceedings of the 62nd Annual Meeting of the
  Association for Computational Linguistics}.
\newblock (2024).

\bibitem{chomsky2002syntactic}
N Chomsky, {\em Syntactic structures}.
\newblock (Mouton de Gruyter), (2002).

\bibitem{chomsky2014aspects}
N Chomsky, {\em Aspects of the Theory of Syntax}.
\newblock (MIT press), (2014).

\bibitem{nevins2009piraha}
A Nevins, D Pesetsky, C Rodrigues, Pirah{\~a} exceptionality: A reassessment.
\newblock {\em\protect\JournalTitle{Language}} (2009).

\bibitem{kotei1972west}
S Kotei, {\em The West African autochtonous alphabets: an exercise in
  comparative palaeography}.
\newblock (na), (1972).

\bibitem{willis1998adinkra}
WB Willis, {\em The Adinkra dictionary: A visual primer on the language of
  Adinkra}.
\newblock (Pyramid Complex), (1998).

\bibitem{wang2024comparative}
L Wang, R Ball, H Xu, Comparative study of sand drawings in oceania and africa.
\newblock {\em\protect\JournalTitle{arXiv preprint arXiv:2404.04798}} (2024).

\bibitem{grattafiori2024llama}
A Grattafiori, et~al., The llama 3 herd of models.
\newblock {\em\protect\JournalTitle{arXiv preprint arXiv:2407.21783}} (2024).

\bibitem{liu2024spinquant}
Z Liu, et~al., Spinquant: Llm quantization with learned rotations.
\newblock {\em\protect\JournalTitle{arXiv preprint arXiv:2405.16406}} (2024).

\bibitem{yang2024qwen2}
A Yang, et~al., Qwen2.5 technical report.
\newblock {\em\protect\JournalTitle{arXiv preprint arXiv:2412.15115}} (2024).

\bibitem{warstadt2023call}
A Warstadt, et~al., Call for papers--the babylm challenge: Sample-efficient
  pretraining on a developmentally plausible corpus.
\newblock {\em\protect\JournalTitle{arXiv preprint arXiv:2301.11796}} (2023).

\bibitem{kaplan2020scaling}
J Kaplan, et~al., Scaling laws for neural language models.
\newblock {\em\protect\JournalTitle{arXiv preprint arXiv:2001.08361}} (2020).

\bibitem{jawahar2019does}
G Jawahar, B Sagot, D Seddah, What does bert learn about the structure of
  language? in {\em ACL 2019-57th Annual Meeting of the Association for
  Computational Linguistics}.
\newblock (2019).

\bibitem{tenney2019you}
I Tenney, et~al., What do you learn from context? probing for sentence
  structure in contextualized word representations.
\newblock {\em\protect\JournalTitle{arXiv preprint arXiv:1905.06316}} (2019).

\bibitem{raffel2020exploring}
C Raffel, et~al., Exploring the limits of transfer learning with a unified
  text-to-text transformer.
\newblock {\em\protect\JournalTitle{Journal of machine learning research}}
  (2020).

\bibitem{geva2020transformer}
M Geva, R Schuster, J Berant, O Levy, Transformer feed-forward layers are
  key-value memories.
\newblock {\em\protect\JournalTitle{arXiv preprint arXiv:2012.14913}} (2020).

\bibitem{clark2019does}
K Clark, U Khandelwal, O Levy, CD Manning, What does bert look at? an analysis
  of bert's attention.
\newblock {\em\protect\JournalTitle{arXiv preprint arXiv:1906.04341}} (2019).

\bibitem{hewitt2019structural}
J Hewitt, CD Manning, A structural probe for finding syntax in word
  representations in {\em Proceedings of the 2019 Conference of the North
  American Chapter of the Association for Computational Linguistics: Human
  Language Technologies, Volume 1 (Long and Short Papers)}.
\newblock (2019).

\bibitem{tenney2019bert}
I Tenney, D Das, E Pavlick, Bert rediscovers the classical nlp pipeline.
\newblock {\em\protect\JournalTitle{arXiv preprint arXiv:1905.05950}} (2019).

\bibitem{voita2019bottom}
E Voita, R Sennrich, I Titov, The bottom-up evolution of representations in the
  transformer: A study with machine translation and language modeling
  objectives.
\newblock {\em\protect\JournalTitle{arXiv preprint arXiv:1909.01380}} (2019).

\bibitem{NIPS2017_3f5ee243}
A Vaswani, et~al., Attention is all you need in {\em Advances in Neural
  Information Processing Systems}.
\newblock (2017).

\bibitem{voita2019analyzing}
E Voita, D Talbot, F Moiseev, R Sennrich, I Titov, Analyzing multi-head
  self-attention: Specialized heads do the heavy lifting, the rest can be
  pruned.
\newblock {\em\protect\JournalTitle{arXiv preprint arXiv:1905.09418}} (2019).

\bibitem{rogers2021primer}
A Rogers, O Kovaleva, A Rumshisky, A primer in bertology: What we know about
  how bert works.
\newblock {\em\protect\JournalTitle{Transactions of the association for
  computational linguistics}} (2021).

\bibitem{fabiani2022neural}
F Fabiani, PJ Goulart, Neural network controllers for uncertain linear systems.
\newblock {\em\protect\JournalTitle{arXiv preprint arXiv:2204.13209}} (2022).

\bibitem{sun2024achieving}
A Sun, C Ma, K Ge, S Vosoughi, Achieving domain-independent certified
  robustness via knowledge continuity.
\newblock {\em\protect\JournalTitle{arXiv preprint arXiv:2411.01644}} (2024).

\bibitem{huster2019limitations}
T Huster, CYJ Chiang, R Chadha, Limitations of the lipschitz constant as a
  defense against adversarial examples in {\em ECML PKDD 2018 Workshops:
  Nemesis 2018, UrbReas 2018, SoGood 2018, IWAISe 2018, and Green Data Mining
  2018, Dublin, Ireland, September 10-14, 2018, Proceedings 18}.
\newblock (2019).

\bibitem{jia2024aligning}
Y Jia, C Zhang, S Vosoughi, Aligning relational learning with lipschitz
  fairness in {\em The Twelfth International Conference on Learning
  Representations}.
\newblock (2024).

\bibitem{yang2022tableformer}
J Yang, et~al., Tableformer: Robust transformer modeling for table-text
  encoding in {\em Proceedings of the 60th Annual Meeting of the Association
  for Computational Linguistics}.
\newblock (2022).

\bibitem{verine2023expressivity}
A Verine, B Negrevergne, Y Chevaleyre, F Rossi, On the expressivity of
  bi-lipschitz normalizing flows in {\em Asian Conference on Machine Learning}.
\newblock (2023).

\bibitem{gupta2022multivariate}
K Gupta, F Kaakai, B Pesquet-Popescu, JC Pesquet, FD Malliaros, Multivariate
  lipschitz analysis of the stability of neural networks.
\newblock {\em\protect\JournalTitle{Frontiers in Signal Processing}} (2022).

\bibitem{jia2023enhancing}
Y Jia, D Zou, H Wang, H Jin, Enhancing node-level adversarial defenses by
  lipschitz regularization of graph neural networks in {\em Proceedings of the
  29th ACM SIGKDD Conference on Knowledge Discovery and Data Mining}.
\newblock (2023).

\bibitem{gama2019stability}
F Gama, A Ribeiro, J Bruna, Stability of graph scattering transforms.
\newblock {\em\protect\JournalTitle{Advances in Neural Information Processing
  Systems}} (2019).

\bibitem{ruan2018reachability}
W Ruan, X Huang, M Kwiatkowska, Reachability analysis of deep neural networks
  with provable guarantees.
\newblock {\em\protect\JournalTitle{arXiv preprint arXiv:1805.02242}} (2018).

\bibitem{sokolic2017robust}
J Sokoli{\'c}, R Giryes, G Sapiro, MR Rodrigues, Robust large margin deep
  neural networks.
\newblock {\em\protect\JournalTitle{IEEE Transactions on Signal Processing}}
  (2017).

\bibitem{behrmann2019invertible}
J Behrmann, W Grathwohl, RT Chen, D Duvenaud, JH Jacobsen, Invertible residual
  networks in {\em International conference on machine learning}.
\newblock (2019).

\bibitem{berwick2016only}
RC Berwick, N Chomsky, {\em Why only us: Language and evolution}.
\newblock (MIT press), (2016).

\bibitem{hauser2002faculty}
MD Hauser, N Chomsky, WT Fitch, The faculty of language: what is it, who has
  it, and how did it evolve?
\newblock {\em\protect\JournalTitle{science}} (2002).

\bibitem{pinker2003language}
S Pinker, {\em The language instinct: How the mind creates language}.
\newblock (Penguin uK), (2003).

\bibitem{lidz2003infants}
J Lidz, S Waxman, J Freedman, What infants know about syntax but couldn't have
  learned: Experimental evidence for syntactic structure at 18 months.
\newblock {\em\protect\JournalTitle{Cognition}} (2003).

\bibitem{linzen2016assessing}
T Linzen, E Dupoux, Y Goldberg, Assessing the ability of lstms to learn
  syntax-sensitive dependencies.
\newblock {\em\protect\JournalTitle{Transactions of the Association for
  Computational Linguistics}} (2016).

\bibitem{goldberg2019assessing}
Y Goldberg, Assessing bert's syntactic abilities.
\newblock {\em\protect\JournalTitle{arXiv preprint arXiv:1901.05287}} (2019).

\bibitem{hupkes2020compositionality}
D Hupkes, V Dankers, M Mul, E Bruni, Compositionality decomposed: How do neural
  networks generalise?
\newblock {\em\protect\JournalTitle{Journal of Artificial Intelligence
  Research}} (2020).

\bibitem{warstadt2020blimp}
A Warstadt, et~al., Blimp: The benchmark of linguistic minimal pairs for
  english.
\newblock {\em\protect\JournalTitle{Transactions of the Association for
  Computational Linguistics}} (2020).

\bibitem{bogdan2025emergent}
PC Bogdan, Emergent effects of scaling on the functional hierarchies within
  large language models.
\newblock {\em\protect\JournalTitle{arXiv preprint arXiv:2501.07359}} (2025).

\bibitem{hu2025between}
MY Hu, J Petty, C Shi, W Merrill, T Linzen, Between circuits and chomsky:
  Pre-pretraining on formal languages imparts linguistic biases.
\newblock {\em\protect\JournalTitle{arXiv preprint arXiv:2502.19249}} (2025).

\bibitem{mikolov2013distributed}
T Mikolov, I Sutskever, K Chen, GS Corrado, J Dean, Distributed representations
  of words and phrases and their compositionality in {\em Advances in neural
  information processing systems}.
\newblock (2013).

\bibitem{ilic2018deep}
S Ili{\'c}, E Marrese-Taylor, JA Balazs, Y Matsuo, Deep contextualized word
  representations for detecting sarcasm and irony.
\newblock {\em\protect\JournalTitle{arXiv preprint arXiv:1809.09795}} (2018).

\bibitem{petroni2019language}
F Petroni, et~al., Language models as knowledge bases?
\newblock {\em\protect\JournalTitle{arXiv preprint arXiv:1909.01066}} (2019).

\bibitem{malik2025constructed}
S Malik-Moraleda, et~al., Constructed languages are processed by the same brain
  mechanisms as natural languages.
\newblock {\em\protect\JournalTitle{Proceedings of the National Academy of
  Sciences}} (2025).

\bibitem{Ghorbanpour2022impossiblelanguages}
A Ghorbanpour, Book review: Impossible languages.
\newblock {\em\protect\JournalTitle{Journal of Universal Language}} (2022).

\bibitem{chowdhery2023palm}
A Chowdhery, et~al., Palm: Scaling language modeling with pathways.
\newblock {\em\protect\JournalTitle{Journal of Machine Learning Research}}
  (2023).

\end{thebibliography}


\begin{thebibliography}{1}

\bibitem{kallini2024mission}
J Kallini, I Papadimitriou, R Futrell, K Mahowald, C Potts, Mission: Impossible
  language models in {\em Proceedings of the 62nd Annual Meeting of the
  Association for Computational Linguistics (Volume 1: Long Papers)}.
\newblock (2024).

\bibitem{loshchilov2017decoupled}
I Loshchilov, Decoupled weight decay regularization.
\newblock {\em\protect\JournalTitle{arXiv preprint arXiv:1711.05101}} (2017).

\bibitem{tian2024fast}
J Tian, YC Liu, JS Smith, Z Kira, Fast trainable projection for robust
  fine-tuning.
\newblock {\em\protect\JournalTitle{Advances in Neural Information Processing
  Systems}} (2024).

\end{thebibliography}

\end{document}



\maketitle

\SItext


\section*{Supplemental Theory}
\subsection{Impossible Language Construction and Classification}
\leavevmode  

\subsubsection{Impossible Language Examples}
\leavevmode 

We conduct a series of experiments based on impossible languages. For clarity, we provide two sentence examples and their corresponding variants across \textbf{Shuffle}, \textbf{Reverse}, and \textbf{Hop} language types in Table~\ref{tab:impossible-languages-examples}. These examples refer to Kallini et al.~\cite{kallini2024mission}, providing a clear approach for constructing impossible languages. 

\begin{table*}[!htbp]
  \centering
  \caption{Examples of impossible languages across three perturbation types: \textbf{Shuffle}, \textbf{Reverse}, and \textbf{Hop}. Different colors highlight distinct sentence components. \tokensing\ and \tokenrev\ denote markers in \textbf{Reverse} and \textbf{Hop} settings, respectively.}
  \label{tab:impossible-languages-examples}
  \resizebox{1.0\textwidth}{!}{%
  \begin{tabular}{l|l|l|l}
    \toprule
    \textbf{\textbf{Class}} & \textbf{Language} & \textbf{Example 1} & \textbf{Example 2} \\ 
    \midrule
    \multirow{10}{*}{\textbf{Shuffle}} 
    & \texttt{NoShuffle} & 
    \texttt{\tokenHe\,\tokencleans\,\tokenhis\,\tokenvery\,\tokenmessy\,\tokenbooks\,\tokenhe\,\tokenlf\,\tokenperiod} & 
    \texttt{\tokenThey\,\tokenclean\,\tokenhis\,\tokenvery\,\tokenmessy\,\tokenbooks\,\tokenhe\,\tokenlf\,\tokenperiod} \\[3pt] 
    \cline{2-4}
    & \texttt{NondeterministicShuffle} & 
    \texttt{\tokenmessy\,\tokenbooks\,\tokenhis\,\tokenhe\,\tokenvery\,\tokenperiod\,\tokenlf\,\tokenHe\,\tokencleans} & 
    \texttt{\tokenhis\,\tokenperiod\,\tokenvery\,\tokenhe\,\tokenThey\,\tokenmessy\,\tokenlf\,\tokenbooks\,\tokenclean} \\[3pt] 
    \cline{2-4}
    & \texttt{DeterministicShuffle(s=21)} & 
    \texttt{\tokencleans\,\tokenHe\,\tokenmessy\,\tokenbooks\,\tokenhe\,\tokenlf\,\tokenvery\,\tokenperiod\,\tokenhis} & 
    \texttt{\tokenclean\,\tokenThey\,\tokenmessy\,\tokenbooks\,\tokenhe\,\tokenlf\,\tokenvery\,\tokenperiod\,\tokenhis} \\[3pt] 
    \cline{2-4}
    & \texttt{DeterministicShuffle(s=57)} & 
    \texttt{\tokencleans\,\tokenhis\,\tokenHe\,\tokenmessy\,\tokenperiod\,\tokenhe\,\tokenvery\,\tokenlf\,\tokenbooks} & 
    \texttt{\tokenclean\,\tokenhis\,\tokenThey\,\tokenmessy\,\tokenperiod\,\tokenhe\,\tokenvery\,\tokenlf\,\tokenbooks} \\[3pt] 
    \cline{2-4}
    & \texttt{DeterministicShuffle(s=84)} & 
    \texttt{\tokenHe\,\tokenmessy\,\tokenperiod\,\tokenlf\,\tokenhis\,\tokenvery\,\tokenbooks\,\tokencleans\,\tokenhe} & 
    \texttt{\tokenThey\,\tokenmessy\,\tokenperiod\,\tokenlf\,\tokenhis\,\tokenvery\,\tokenbooks\,\tokenclean\,\tokenhe} \\[3pt] 
    \cline{2-4}
    & \texttt{LocalShuffle(w=3)} & 
    \texttt{\tokenhis\,\tokenHe\,\tokencleans\,\tokenbooks\,\tokenvery\,\tokenmessy\,\tokenperiod\,\tokenhe\,\tokenlf} & 
    \texttt{\tokenhis\,\tokenThey\,\tokenclean\,\tokenbooks\,\tokenvery\,\tokenmessy\,\tokenperiod\,\tokenhe\,\tokenlf} \\[3pt] 
    \cline{2-4}
    & \texttt{LocalShuffle(w=5)} & 
    \texttt{\tokenhis\,\tokenmessy\,\tokenvery\,\tokenHe\,\tokencleans\,\tokenlf\,\tokenbooks\,\tokenhe\,\tokenperiod} & 
    \texttt{\tokenhis\,\tokenmessy\,\tokenvery\,\tokenThey\,\tokenclean\,\tokenlf\,\tokenbooks\,\tokenhe\,\tokenperiod} \\[3pt] 
    \cline{2-4}
    & \texttt{LocalShuffle(w=10)} & 
    \texttt{\tokenmessy\,\tokenbooks\,\tokenhis\,\tokenhe\,\tokenvery\,\tokenperiod\,\tokenlf\,\tokenHe\,\tokencleans} & 
    \texttt{\tokenmessy\,\tokenbooks\,\tokenhis\,\tokenhe\,\tokenvery\,\tokenperiod\,\tokenlf\,\tokenThey\,\tokenclean} \\[3pt] 
    \cline{2-4}
    & \texttt{EvenOddShuffle} & 
    \texttt{\tokenHe\,\tokenhis\,\tokenmessy\,\tokenhe\,\tokenperiod\,\tokencleans\,\tokenvery\,\tokenbooks\,\tokenlf} & 
    \texttt{\tokenThey\,\tokenhis\,\tokenmessy\,\tokenhe\,\tokenperiod\,\tokenclean\,\tokenvery\,\tokenbooks\,\tokenlf} \\[3pt]
    \midrule
    
    \multirow{3}{*}{\textbf{Reverse}}
    & \texttt{NoReverse} & 
    \texttt{\tokenHe\,\tokencleans\,\tokenhis\,\tokenvery\,\tokenmessy\,\tokenbooks\,\tokenrev\,\tokenhe\,\tokenlf\,\tokenperiod} & 
    \texttt{\tokenThey\,\tokenclean\,\tokenhis\,\tokenrev\,\tokenvery\,\tokenmessy\,\tokenbooks\,\tokenhe\,\tokenlf\,\tokenperiod} \\[3pt] 
    \cline{2-4}
    & \texttt{PartialReverse} & 
    \texttt{\tokenHe\,\tokencleans\,\tokenhis\,\tokenvery\,\tokenmessy\,\tokenbooks\,\tokenrev\,\tokenperiod\,\tokenlf\,\tokenhe} & 
    \texttt{\tokenThey\,\tokenclean\,\tokenhis\,\tokenrev\,\tokenperiod\,\tokenlf\,\tokenhe\,\tokenbooks\,\tokenmessy\,\tokenvery} \\[3pt] 
    \cline{2-4}
    & \texttt{FullReverse} & 
    \texttt{\tokenperiod\,\tokenlf\,\tokenhe\,\tokenrev\,\tokenbooks\,\tokenmessy\,\tokenvery\,\tokenhis\,\tokencleans\,\tokenHe} & 
    \texttt{\tokenperiod\,\tokenlf\,\tokenhe\,\tokenbooks\,\tokenmessy\,\tokenvery\,\tokenrev\,\tokenhis\,\tokenclean\,\tokenThey} \\[3pt]
    \midrule
    
    \multirow{3}{*}{\textbf{Hop}}
    & \texttt{NoHop} & 
    \texttt{\tokenHe\,\tokenclean\,\tokensing\,\tokenhis\,\tokenvery\,\tokenmessy\,\tokenbooks\,\tokenhe\,\tokenlf\,\tokenperiod} &  
    \texttt{\tokenThey\,\tokenclean\,\tokenplur\,\tokenhis\,\tokenvery\,\tokenmessy\,\tokenbooks\,\tokenhe\,\tokenlf\,\tokenperiod} \\[3pt] 
    \cline{2-4}
    & \texttt{Tokens4Hop} & 
    \texttt{\tokenHe\,\tokenclean\,\tokenhis\,\tokenvery\,\tokenmessy\,\tokenbooks\,\tokensing\,\tokenhe\,\tokenlf\,\tokenperiod} &  
    \texttt{\tokenThey\,\tokenclean\,\tokenhis\,\tokenvery\,\tokenmessy\,\tokenbooks\,\tokenplur\,\tokenhe\,\tokenlf\,\tokenperiod} \\[3pt] 
    \cline{2-4}
    & \texttt{Words4Hop} & 
    \texttt{\tokenHe\,\tokenclean\,\tokenhis\,\tokenvery\,\tokenmessy\,\tokenbooks\,\tokenhe\,\tokenlf\,\tokensing\,\tokenperiod} &  
    \texttt{\tokenThey\,\tokenclean\,\tokenhis\,\tokenvery\,\tokenmessy\,\tokenbooks\,\tokenhe\,\tokenlf\,\tokenplur\,\tokenperiod} \\[3pt]
    \bottomrule
  \end{tabular}
  }
\end{table*}

\subsubsection{Unconventional Languages and Symbol Systems Beyond Grammar}
\leavevmode  

\begin{figure*}[h]
\label{uncon_languages}
\begin{center}
\includegraphics[width=0.9\linewidth]{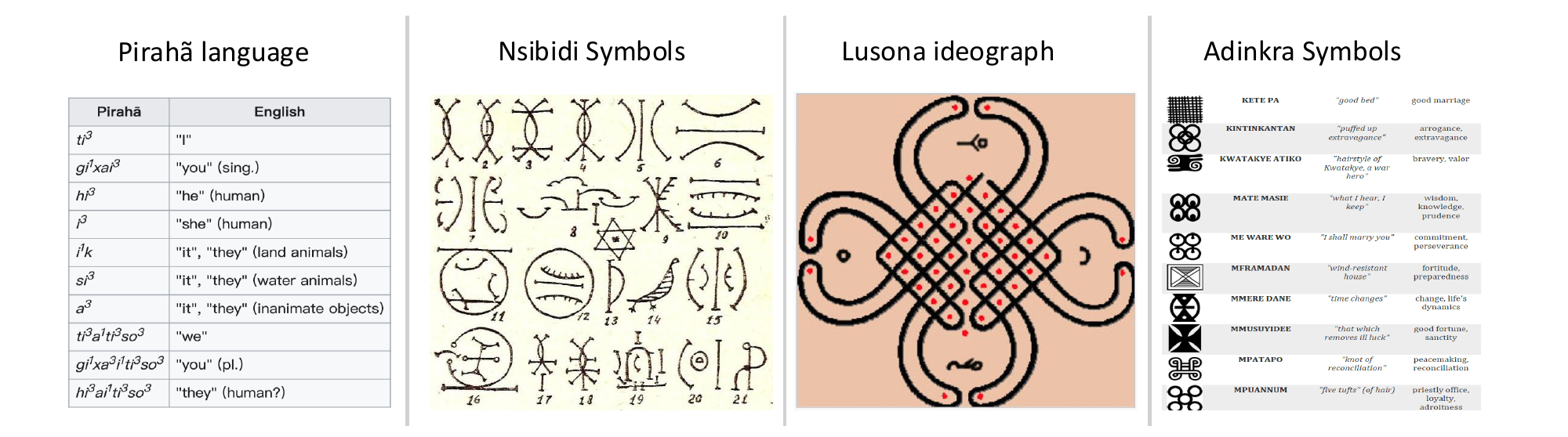}
\end{center}
\caption{These examples compare various unconventional linguistic and symbolic systems that emphasize semantic meaning over syntax rules. From left to right, it features: \textbf{Pirahã Language} (source: \url{https://images.app.goo.gl/rAtsj2G61iH6DDWs8}), \textbf{Nsibidi Symbols} (source: \url{https://images.app.goo.gl/Abvv764nE3MrB4HH6}), \textbf{Lusona Ideographs} (source: \url{https://images.app.goo.gl/T382r9TNi217LMqs7}), and \textbf{Adinkra Symbols} (source: \url{www.kentecloth.net/kente-cloth-adinkra-symbols-meaning}).}
\vspace{0.2cm}
\label{fig:unconventional-languages}
\end{figure*}

We present several unconventional linguistic and symbolic systems, as shown in Fig.~\ref{fig:unconventional-languages}, including four types:
\begin{itemize}
    \item \textbf{Pirahã Language:} A table showcasing pronouns in the Pirahã language, spoken by the Pirahã people of the Amazon. It is notable for its lack of recursion and unique approach to linguistic expression, emphasizing direct experience rather than complex grammatical structures.
    \item \textbf{Nsibidi Symbols:} An ideographic writing system from southeastern Nigeria and Cameroon, historically used by the Ekpe secret society and other groups to convey complex ideas, emotions, and social status. Nsibidi symbols function as a visual language rather than a conventional alphabet.
    \item \textbf{Lusona Ideographs:} A collection of symbolic designs originating from the Lunda and Chokwe people of Central Africa. These ideographs are traditionally drawn in the sand and serve as mnemonic devices, conveying oral stories, philosophical concepts, and mathematical ideas.
    \item \textbf{Adinkra Symbols:} A set of visual symbols from the Akan people of Ghana, used to represent proverbs, historical events, and moral values. Each symbol carries deep cultural significance and is commonly found in textiles, pottery, and architectural designs.
\end{itemize}

These linguistic systems prioritize semantics over syntactic rules, meaning that semantic meaning plays a key role in communication rather than syntactic rules. This further supports our argument: a language can be learned by language models (LMs) as long as it possesses complete semantic. Scaling LMs enhances their semantic understanding, enabling them to learn even so-called "impossible" languages.

\subsubsection{Sematic Coherence Explanation}
\leavevmode 

We introduce \textbf{semantic coherence (SC)} in our paper as a measure of how effectively semantic information is conveyed through language structures. SC is categorized into three levels: $0$, $-1$, and $-2$, which we refer to as \textit{SC(0), SC(-1), and SC(-2)}. Each level represents a different degree of syntactic and semantic retention:

\begin{itemize}  
    \item \textit{SC(0)}: Languages that fully preserve both syntactic and semantic structures.  
    \item \textit{SC(-1)}: Languages that retain complete semantic information but exhibit partial syntactic loss.  
    \item \textit{SC(-2)}: Languages that have incomplete retention of both syntactic and semantic structures.  
\end{itemize}

The retention of syntactic information in a language depends on whether the natural language syntax has been perturbed. If any perturbation is applied, the language is considered to have incomplete syntactic retention. Therefore, except for natural languages, which serve as control groups, all other languages exhibit incomplete syntactic retention. Similarly, for semantic information retention, if the perturbation function is deterministic, the perturbed language can be fully recovered into a natural language. In this case, we say the language has complete semantic retention. Otherwise, if recovery is not possible, the language has incomplete semantic retention. Among all perturbation functions, only \texttt{NondeterministicShuffle} is entirely random, making the corresponding languages unrecoverable. Therefore, it is the only function that results in incomplete semantic retention.

\subsection{Lipschitz Perspective on LLM Stability}
\leavevmode  

\textit{Why does the Lipschitz constraint improve the performance of LLMs in learning impossible languages?} The Lipschitz-optimizer is designed with this principle in mind. Here we explore this question, starting with the formal definition of the Lipschitz constant.

A function $f:\mathbb{R}^{n} \rightarrow \mathbb{R}^{m}$ is considered to be Lipschitz continuous on an input set $\mathcal{X}\subseteq  \mathbb{R}^{n}$ if there exists a bound $K \geq 0$ such that for all $\vx,\vy \in \mathcal{X}$, $f$ satisfies:
\vspace{-0.3em}
\begin{equation}
\label{eq-1}
    \norm{f(\vx) - f(\vy)} \le K\norm{\vx-\vy}, \forall \vx,\vy \in \mathcal{X}.
\end{equation}
The smallest possible $K$ in Equation~(\ref{eq-1}) is referred to as the Lipschitz constant of $f$. When f represents a neural network, \textit{the K essentially quantifies the maximum change in the output space corresponding to a variation in the input space.} It reflects the network's sensitivity to variations or perturbations in its input features, making the Lipschitz constant an important metric for evaluating the stability of a neural network.

\subsubsection{Stability for Fine-Tuned models}
\leavevmode

We analyze the stability of fully fine-tuned LLMs through the lens of Lipschitz continuity, focusing on how controlled Lipschitz constants mitigate sensitivity to input perturbations. Our analysis begins with foundational definitions and builds to actionable stability guarantees.

Let the input space $ \mathcal{X} \subseteq \mathbb{R}^m$ be a compact set equipped with norm $ \lVert \cdot \rVert_i$, and let $ g: \mathbb{R}^m \to \mathbb{R}^n$ denote a mapping function with output norm $ \lVert \cdot \rVert_o$. Given two input vectors $ \vx_1, \vx_2 \in \mathcal{X}$, we define the fine-tuning difference as $ \Delta g(\cdot) := g_f(\cdot) - g_o(\cdot)$, where $ g_o(\cdot)$ and $ g_f(\cdot)$ represent the pre-trained and fine-tuned models respectively. 

Three critical assumptions ground our analysis: (i) The pre-trained model $ g_o$ and fine-tuned model $ g_f$ are both Lipschitz continuous over the input space $ \mathcal{X}$, with respective constants $ L_o$ and $ L_f$; 
(ii) All activation functions (e.g., ReLU, softmax) in $ g_o$ and $ g_f$ are Lipschitz continuous, excluding non-smooth operations; 
(iii) The input space $ \mathcal{X} \subseteq \mathbb{R}^m$ is compact, ensuring the existence of global Lipschitz bounds for compositional mappings.
Under these assumptions, the fine-tuning difference $ \Delta g(\cdot)$ inherits Lipschitz continuity with constant $ L_g$. This leads to two fundamental inequalities:
\begin{equation}
\label{eq:delta-bound}
\lVert \Delta g(\vx_1) - \Delta g(\vx_2) \rVert_o \leq L_g \lVert \vx_1 - \vx_2 \rVert_i,
\end{equation}
\begin{equation}
\label{eq:pretrained-bound}
\lVert g_o(\vx_1) - g_o(\vx_2) \rVert_o \leq L_o \lVert \vx_1 - \vx_2 \rVert_i.
\end{equation}
Combining these through the triangle inequality reveals the composite Lipschitz bound governing stability:
\begin{equation}
\label{eq:composite-bound}
\begin{aligned}
\lVert g_f(\vx_1) - g_f(\vx_2) \rVert_o &\leq \lVert g_o(\vx_1) - g_o(\vx_2) \rVert_o + \lVert \Delta g(\vx_1) - \Delta g(\vx_2) \rVert_o \\
&\leq (L_o + L_g) \lVert \vx_1 - \vx_2 \rVert_i.
\end{aligned}
\end{equation}
Thus, the stability of the fine-tuned model depends on the combined Lipschitz constant $ L_o + L_g$. During full fine-tuning, minimizing $ L_g$ (the Lipschitz constant of the fine-tuning difference) directly enhances stability, as $ L_o$ remains fixed for the pre-trained model.


\subsubsection{Stability for Impossible Languages}
\leavevmode  

Impossible languages can be considered as natural languages augmented with perturbations, which aligns with the definition of impossible languages presented in this paper. Let $ \vx_2$ denote natural languages, and $ \vx_1 = \vx_2 + \delta$ represent impossible languages. Then, Eq.\ref{eq:composite-bound} can be expressed as:
\vspace{-0.5em}
\begin{equation}
\label{eq-4}
   \|g_f(\vx_1) - g_f(\vx_2)\|_o \le (L_o+ L_g) \|\delta\|_i,
\end{equation}
where $ \delta$ represents the imposed perturbation on natural languages, with a larger $ \delta$ indicating a higher position on the impossibility continuum. From this, it follows that the Lipschitz constant $ L_d + L_o$ and perturbation are closely related to the stability of LLMs. When $ \delta$ is fixed, a smaller Lipschitz constant implies that the model is less influenced by perturbations in the output space, thereby exhibiting better performance and stability.

\subsubsection{Correlation Between Linear Module Stability and LLM Stability} 
\leavevmode  
\label{ssec:linear-module-correlation}

The compositional nature of LLMs allows us to characterize their stability through a Lipschitz analysis of constituent modules. While direct computation of exact Lipschitz constants for full LLMs remains computationally prohibitive, we establish an upper bound by the following Lemma.

\begin{lemma}
Consider an LLM composed of $m\) linear layers $g_i(x) = W_i x + b_i\) and $n\) nonlinear activation functions $\sigma_j\). The global Lipschitz constant of this model satisfies:
\begin{equation}
\label{eq:lip-bound}
L_{\text{LLM}} \leq \left(\prod_{i=1}^m \|W_i\|_2\right)\left(\prod_{j=1}^n L_{\sigma_j}\right),
\end{equation}
where $\|W_i\|_2\) denotes the spectral norm of the weight matrix $W_i\), and the activation Lipschitz constants $L_{\sigma_j}\leq 1\) for standard activations (e.g., ReLU: $L_{\sigma}=1\), softmax with temperature $\tau\): $L_{\sigma}=1/\tau\)).
\end{lemma}

\begin{proof}
Let $F(x)$ be the composite function representing the LLM:
\begin{equation}
F(x) = (\sigma_m \circ g_n \circ \cdots \circ \sigma_1 \circ g_1)(x),
\end{equation}
where each $g_i\) is a linear transformation and each $\sigma_j\) is a nonlinear activation function. The bound (Eq.~\ref{eq:lip-bound}) follows directly from the sequential application of Lipschitz continuity properties: first, each linear layer $g_i\) amplifies the norm of differences by at most its spectral norm $\|W_i\|_2\); second, nonlinear activations $\sigma_j\) are standard and thus non-expansive ($L_{\sigma_j}\leq 1\)); third, residual connections, if present, add perturbations linearly rather than multiplicatively, thus not increasing the Lipschitz constant beyond multiplicative bounds already established.
By iteratively applying these properties, we have:
\begin{equation}
\|F(x)-F(y)\| \leq \|W_n\|_2 L_{\sigma_{n-1}}\|W_{n-1}\|_2 \cdots L_{\sigma_1}\|W_1\|_2 \|x-y\|.
\end{equation}
Hence, Eq.~\ref{eq:lip-bound} holds directly.
\end{proof}

\textbf{Spectral Dominance Mechanism.} The Lipschitz constant of the linear module is given by the spectral norm of its weight matrix.
For example, for linear module $g_i(x) = W_ix + b_i$, its Lipschitz constant can be obtained as follow: 
\vspace{-0.5em}
\begin{equation}
\label{eq:spectral_norm}
L_{g_i} = \|W_i\|_2 = \sigma_{max} = \sqrt{\lambda_{\text{max}}(W_i^T W_i)},
\end{equation}
where $\sigma_{max}$ is max sigular value of $W$ and $\lambda_{\text{max}} $ denotes the largest eigenvalue of $ W_i^T W_i $. Combining Eq.~\ref{eq:lip-bound} and Eq.~\ref{eq:spectral_norm}, we conclude that during the full fine-tuning process, constraining the Lipschitz constants of linear modules in LLMs can significantly enhance their stability.

\subsection{Spectral-Constrained Optimization for Stable Training}
\leavevmode  

The training of modern LLMs often encounters instability issues due to the presence of unstable gradients. To tackle this, we propose various methods for controlling spectral norms, which provide mathematical guarantees for stability.

\subsubsection{Initial Update} 
\leavevmode  

The first step of the optimization updates the weights via adaptive stochastic optimization. For $ W_{t-1} \in \mathbb{R}^{m \times n}$, the $t$-th iteration is:
\begin{equation}
\label{eq-15}
W_t \leftarrow \Phi\left(W_{t-1}, \nabla_{W_{t-1}} \mathcal{L}_{\mathcal{D}_{tr}}(W_{t-1})\right),
\end{equation}
where $\mathcal{L}_{\mathcal{D}_{tr}}$ is the mini-batch loss, and $\Phi(\cdot)$ denotes the AdamW optimizer \cite{loshchilov2017decoupled} with decoupled weight decay.


\subsubsection{Spectral Norm Clipping}
\leavevmode  
\label{sssec:spectral-clipping}
For a linear module's weight matrix $W \in \mathbb{R}^{m \times n}$, we perform singular value decomposition (SVD):
\begin{equation}
W = U\Sigma V^\top,\quad 
\begin{cases} 
U \in \mathbb{R}^{m \times m} & \text{(orthogonal)} \\
V \in \mathbb{R}^{n \times n} & \text{(orthogonal)} \\
\Sigma = \operatorname{diag}(\sigma_1, \ldots, \sigma_r) & \text{(singular values, } r = \min(m,n)).
\end{cases}
\end{equation}
The spectral norm clipping operation constrains each singular value:
\begin{equation}
\label{eq:clip}
\sigma_i' = \min(\sigma_i, \tau) \quad \text{for } i = 1,\ldots,r.
\end{equation}
Reconstructing the modified weight matrix:
\begin{equation}
W' = U\Sigma'V^\top,\quad \Sigma' = \operatorname{diag}(\sigma_1', \ldots, \sigma_r').
\end{equation}
This guarantees the spectral norm bound:
\begin{equation}
\label{eq:bound}
\|W'\|_2 = \max_{1\leq i\leq r} \sigma_i' \leq \tau,
\end{equation}
where $\|W'\|_2$ denotes the spectral norm. The procedure effectively prevents excessive growth of spectral norms during training while preserving the original parameter structure.

\subsubsection{Frobenius Norm Regularization} 
\leavevmode  
\label{sssec:frobenius-reg}

In addition to spectral norm clipping, we propose Frobenius norm regularization for indirect spectral control. For weight matrix $W \in \mathbb{R}^{m \times n}$, the Frobenius norm is defined as:
\begin{equation}
\label{eq:frobenius-def}
\|W\|_F = \sqrt{\sum_{i=1}^m \sum_{j=1}^n W_{ij}^2} = \sqrt{\sum_{i=1}^r \sigma_i^2} \quad (r = \min(m,n)),
\end{equation}
where $\sigma_i$ are singular values. This establishes the inequality:
\begin{equation}
\|W\|_2 \leq \|W\|_F \leq \sqrt{r}\|W\|_2,
\end{equation}
indicating that Frobenius Norm Regularization inherently constrains spectral norms. The regularized loss function becomes:
\begin{equation}
\label{eq:reg-loss}
\mathcal{L}_{\text{reg}}(f) = \mathcal{L}(f) + \mu \sum_{k=1}^K \|W_k\|_F^2,
\end{equation}
where $\mu$ is the regularization coefficient. The gradient update rule derives as:
\begin{equation}
\label{eq:grad-update}
\frac{\partial \mathcal{L}_{\text{reg}}}{\partial W_k} = \nabla_{W_k}\mathcal{L} + 2\mu W_k.
\end{equation}
Therefore, in optimizer, for $W$, it can be written as during $t$-th update:
\begin{equation}
\label{eq:param-update}
W_{t+1} = W_t - \eta\left(\nabla_{W_t}\mathcal{L} + 2\mu W_t\right),
\end{equation}
where $\eta$ is the learning rate. This formulation enables seamless integration with standard optimizers like SGD and Adam.

\subsubsection{Weight Projection}
\leavevmode  
\label{sssec:weight-projection}

We employ a weight projection mechanism anchored to initialization to stabilize parameter updates while preserving model plasticity. Given initial weight $W_0 \in \mathbb{R}^{m \times n}$ and current weight $W_t$ at step $t$, the projected weight is computed as:
\begin{equation}
\label{eq:projection}
\Pi(W_t, W_0, \gamma_t) = W_0 + \gamma_t \cdot \frac{W_t - W_0}{\|W_t - W_0\|_\infty},
\end{equation}
where $\gamma_t > 0$ controls the projection radius. It ensures the preservation of the directional information of weight updates, limits the magnitude of deviation from the initial values, and maintains the gradient flow through normalized scaling. Together, these factors enhance both the stability and effectiveness of the training process.
\begin{theorem}
\label{thm:spectral-bound}
Let $\widehat{W}_t = \Pi(W_t, W_0, \gamma_t)$ with residual matrix $R_t = W_t - W_0$. Then the spectral norm satisfies:
\begin{equation}
\label{eq:spectral-bound}
\|\widehat{W}_t\|_2 \leq \|W_0\|_2 + \gamma_t \cdot \sqrt{\operatorname{rank}(R_t)}.
\end{equation}
\end{theorem}
\begin{proof}
The bound follows from norm subadditivity and the inequality
$\|A\|_2 \leq \sqrt{\operatorname{rank}(A)}\|A\|_\infty\). Applying this stepwise, we obtain:
\begin{equation}
\begin{aligned}
\|\widehat{W}_t\|_2 &\leq \|W_0\|_2 + \gamma_t \left\|\frac{R_t}{\|R_t\|_\infty}\right\|_2 \\
&\leq \|W_0\|_2 + \gamma_t \sqrt{\operatorname{rank}(R_t)} \cdot \frac{\|R_t\|_2}{\|R_t\|_\infty} \\
&\leq \|W_0\|_2 + \gamma_t \sqrt{\operatorname{rank}(R_t)}.
\end{aligned}
\end{equation}
Hence, the inequality holds as claimed. As shown in Theorem~\ref{thm:spectral-bound}, the projection mechanism enforces control over spectral norms via $\gamma_t$, while automatically adapting to the intrinsic dimensionality of parameter updates via $\operatorname{rank}(R_t)$.
\end{proof}

\subsubsection{Projection Update} 
\leavevmode  
\label{sssec:projection-update}

The Lipschitz constant of the weight matrix $W$ is principally governed by the projection radius $\gamma_t$ in Theorem \ref{thm:spectral-bound}. To adaptively optimize this critical parameter, we formulate $\gamma_t$ as a learnable variable updated via:
\begin{equation}
\label{eq:gamma_dynamics}
\gamma_t \leftarrow \Phi\Big(\gamma_{t-1},\, \nabla_{\gamma}\mathcal{L}\big(W_{t-1}(\gamma)\big)\Big),
\end{equation}
where $\Phi$ denotes the AdamUpdate optimizer \cite{tian2024fast}. The projection update mechanism achieves layer-wise adaptive control of Lipschitz constants through dynamically adjusting the projection radius $\gamma_t$, while preserving gradient directional consistency and reducing computational overhead via decoupled optimization strategies, ultimately achieving an efficient and robust training process.







\FloatBarrier






\bibliography{appendix/pnas-sample}